\documentclass[]{fairmeta}
\usepackage{mathpazo}
\usepackage{tgpagella}
\usepackage{xcolor}
\usepackage{amsmath}
\usepackage{amssymb}
\usepackage{graphicx}
\usepackage{arydshln}
\usepackage{multirow}
\usepackage{booktabs}
\usepackage{marvosym}  
\usepackage{colortbl}
\usepackage{float}
\usepackage{cuted}  
\usepackage{pifont}
\usepackage{pgfplots}
\usepackage{wrapfig}
\usepackage{lineno}

\usepackage{booktabs}    
\usepackage{array}       
\usepackage{amsmath}     
\usepackage{caption}     
\usepackage{multirow}

\usepackage{array}      
\usepackage{subcaption}     
\usepackage{xcolor}     
\usepackage{colortbl}   
\usepackage[most]{tcolorbox}

\newcommand{\tablestyle}[2]{\setlength{\tabcolsep}{#1}\renewcommand{\arraystretch}{#2}\centering\footnotesize}

\newcolumntype{x}[1]{>{\centering\arraybackslash}p{#1pt}}
\newcolumntype{y}[1]{>{\raggedright\arraybackslash}p{#1pt}}
\newcolumntype{z}[1]{>{\raggedleft\arraybackslash}p{#1pt}}

\let\cite\citep
\crefname{figure}{Fig.}{Figs.}
\crefname{table}{Tab.}{Tabs.}
\usepackage{wrapfig}
\usepackage{booktabs}
\usepackage{makecell}   
\usepackage{array}      
\usepackage{caption}    

\title{VecGlypher: Unified Vector Glyph Generation with Language Models}

\author[1,2,*]{Xiaoke Huang}
\author[1]{Bhavul Gauri}
\author[1]{Kam Woh Ng}
\author[1]{Tony Ng}
\author[1]{Mengmeng Xu}
\author[1]{Zhiheng Liu}
\author[1]{Weiming Ren}
\author[1]{Zhaochong An}
\author[1]{Zijian Zhou}
\author[1]{Haonan Qiu}
\author[2]{Yuyin Zhou}
\author[1]{Sen He}
\author[1]{Ziheng Wang}
\author[1]{Tao Xiang}
\author[1]{Xiao Han}

\affiliation[1]{Meta AI}
\affiliation[2]{UC, Santa Cruz}

\contribution[*]{Work done at Meta}

\definecolor{teaser_blue}{RGB}{75,140,204}
\definecolor{teaser_green}{RGB}{69,169,80}

\date{\today}
\metadata[Project Page]{\url{https://xk-huang.github.io/VecGlypher}}

\abstract{

Vector glyphs are the atomic units of digital typography, yet most learning-based pipelines still depend on carefully curated exemplar sheets and raster-to-vector postprocessing, which limits accessibility and editability.
We introduce VecGlypher, a single multimodal language model that generates high-fidelity vector glyphs directly from text descriptions or image exemplars.
Given style prompts or reference glyph images, and a target character, VecGlypher autoregressively emits SVG path tokens, avoiding raster intermediates and producing editable, watertight outlines in one pass. A typography-aware data and training recipe makes this possible: (i) a large-scale continuation stage on 39K noisy Envato fonts to master SVG syntax and long-horizon geometry, followed by (ii) post-training on 2.5K expert-annotated Google Fonts with descriptive tags and exemplars to align language and imagery with geometry; preprocessing normalizes coordinate frames, canonicalizes paths, de-duplicates families, and quantizes coordinates for stable long-sequence decoding. On cross-family OOD evaluation, VecGlypher substantially outperforms both general-purpose LLMs and specialized vector-font baselines for text-only generation, while image-referenced generation reaches a state-of-the-art performance, with marked gains over DeepVecFont-v2 and DualVector. Ablations show that model scale and the two-stage recipe are critical and that absolute-coordinate serialization yields the best geometry. VecGlypher lowers the barrier to font creation by letting users design with words or exemplars, and provides a scalable foundation for future multimodal design tools.

}

\begin{document}

\maketitle

\begin{center}
  \vspace{+5mm}
  \includegraphics[width=1.0\linewidth]{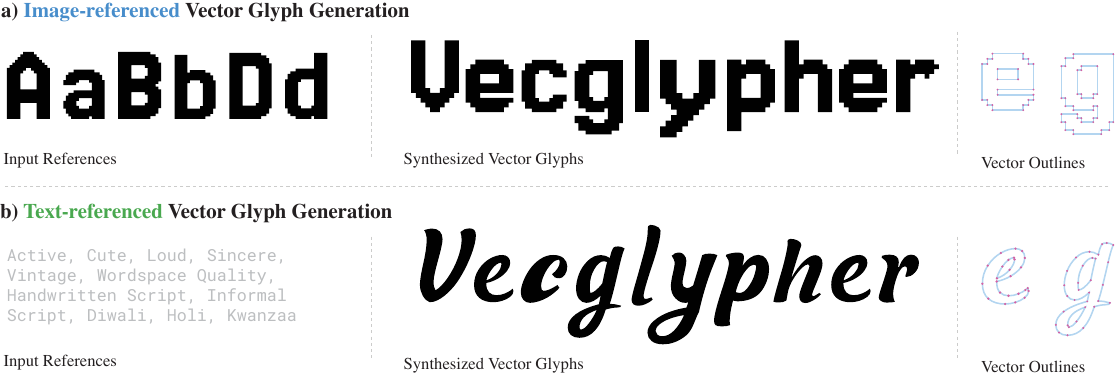}
  \vspace{+0.0mm}
  \captionof{figure}{
\textbf{VecGlypher} generates high-fidelity vector glyphs directly as editable SVG outlines under two types of conditioning: (a) {\textcolor{teaser_blue}{\textbf{image-referenced}}} generation, where a handful of exemplar glyph images specify the style and the model synthesizes new glyphs in the same visual form; and (b) {\textcolor{teaser_green}{\textbf{text-referenced}}} generation, where a natural-language prompt drives the synthesis without requiring exemplars.
The figure shows the synthesized wordmark and sample vector outlines, highlighting one-pass generation of clean, controllable contours for typography workflows.
  }
    \label{fig:teaser-glyph}
\end{center}

\section{Introduction}

Vector glyphs---the parametric curves that define letters and symbols in a font---are the atomic units of digital typography. They must be resolution-independent, compact, and precisely controllable for downstream typesetting, logo design, and UI rendering. Recent learning-based approaches have begun to automate glyph creation, yet the prevailing paradigm remains \textit{image-referenced vector glyph generation}: given a few exemplar raster glyphs of a font, a model synthesizes the remaining characters by predicting vector outlines~\cite{wang2021deepvecfont,wang2023deepvecfontv2,thamizharasan2024vecfusion}. While effective when exemplars are carefully prepared, this workflow assumes that users can already produce or collect representative glyph images for each new style. In practice, this requirement is a major bottleneck, especially for non-experts and for rapid ideation cycles in which designers would rather describe a desired style than draft a reference sheet.

We posit that natural language is a more universal and accessible interface for font creation. Designers and casual users routinely communicate typographic intent with concepts such as \textit{``high-contrast, narrow, slightly condensed, art-deco, playful''}. Moreover, the output we ultimately seek, a sequence of vector drawing commands and coordinates, is itself textual (e.g., an SVG path). This observation suggests an appealing formulation: treat glyph generation as a \textbf{language modeling} problem and leverage multimodal \textbf{Large Language Models} (LLMs)~\cite{bai2025qwen2_5vl,team2025gemma,liu2023visual_llava} to map text descriptions or image exemplars directly to vector code. Besides simplifying the human interface, this formulation stands to inherit the strong text and image understanding of modern foundation models.

However, building an LLM that can actually \textit{draw} credible typography is nontrivial. General-purpose LLMs~\cite{achiam2023gpt4,anthropic2024claude,comanici2025gemini,yang2025qwen3} and off-the-shelf vector-graphics LLMs~\cite{rodriguez2023starvector,xing2025empowering_llm4svg,yang2025omnisvg} that produce icons or simple SVG drawings typically fail on glyphs. Typography imposes strict geometric and stylistic constraints: long sequences of precise coordinates; watertight topology; consistent stroke logic across characters; and subtle style factors (contrast, terminals, serifs, curvature) that must vary coherently with the target content.
Beyond model capacity, practical data issues arise: paired text-glyph examples are scarce~\cite{googlefonts}; tags in large repositories are noisy and often non-visual~\cite{envatoelementsfonts}; and coordinate systems vary across sources. Naively prompting existing LLMs to ``write an SVG path for a serif V'' often yields broken geometry, mismatched case, or degenerate paths.

\begin{wraptable}{r}{0.55\textwidth}
\centering

\caption{
\textbf{SVG draw commands used in this work.} We serialize each glyph as a single SVG path using only \textit{MoveTo}, \textit{LineTo}, \textit{Quadratic Bézier}, and \textit{ClosePath}. Uppercase commands use absolute coordinates; lowercase are relative to the current point. Coordinates are quantized to one decimal; other attributes (e.g., ``fill'', ``stroke'', ``transform'') are unused. 
}

\label{tab:svg_draw_command}
\tablestyle{1pt}{1.0}
{

\begin{tabular}{ccc}
\toprule
\textbf{Command} & \textbf{Arguments} & \textbf{Description} 
\\
\midrule
    \begin{tabular}{c}
        \texttt{M / m} \\
        (MoveTo)
    \end{tabular} &
    $x$, $y$ &
    \begin{tabular}{c}
        Move the cursor to the end-point $(x, y)$ \\
        without drawing anything.
    \end{tabular} 
\\
\midrule
    \begin{tabular}{c}
        \texttt{L / l} \\
        (LineTo)
    \end{tabular} &
    $x$, $y$ & 
    \begin{tabular}{c}
    Draw a line to the end-point $(x, y)$.
    \end{tabular}
\\
\midrule
    \begin{tabular}{c}
        \texttt{Q / q} \\
        (Quadratic \\
        Bézier)
    \end{tabular}
    & 
    \begin{tabular}{c}
            $q_{x}$, $q_{y}$ \\
            $x$, $y$
    \end{tabular}
    & 
    \begin{tabular}{c}
        Draw a quadratic Bézier curve with the control \\
        point $(q_{x}, q_{y})$ and the end-point $(x, y)$.
    \end{tabular}
\\
\midrule
    \begin{tabular}{c}
        \texttt{Z / z} \\ (ClosePath) 
    \end{tabular}
    & 
    $\varnothing$ &
    \begin{tabular}{c}
        Close the path by moving the cursor back \\
        to the starting position (ignorable).
    \end{tabular} 
\\
\bottomrule
\end{tabular}

}
\end{wraptable}

\begin{figure*}[t]
    \centering
    \includegraphics[width=1.0\linewidth]{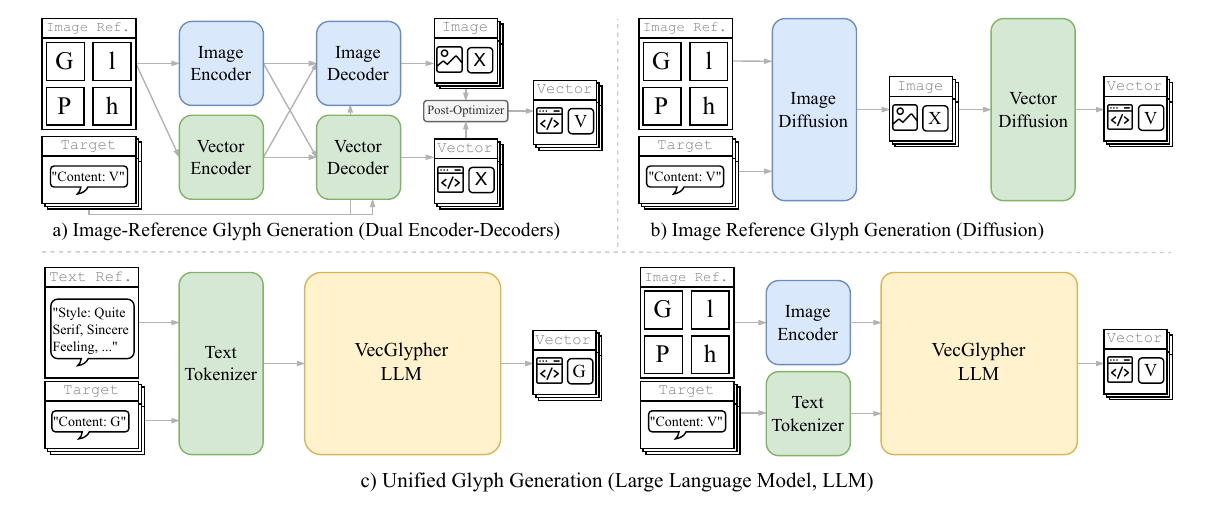}
    \caption{
    \textbf{Paradigm comparisons.} a) Prior image-referenced pipelines use separate image and vector encoder–decoders and a geometry post-optimizer. b) Diffusion-based approaches cascade image diffusion with a vector decoder. c) \textbf{VecGlypher} unifies both text- and image-referenced conditioning within a single LLM: given a style description or reference glyph images plus a target character, the model autoregressively emits SVG path tokens that detokenize to a valid SVG path. This formulation removes raster intermediates and exemplar-sheet requirements while producing directly editable vectors. 
    A \textit{practical workflow} is to first generate a few reference glyphs from text descriptions, then bootstrap with those images to synthesize the full font.
    }
    \label{fig:teaser}
\end{figure*}

\textbf{VecGlypher} addresses these challenges and unifies \textit{text- and image-referenced} vector glyph generation within a single language model (\cref{fig:teaser}). Our model is a multimodal decoder that consumes (i) a style description or image exemplars, and (ii) the target character identity, then autoregressively predicts SVG path tokens for the output glyph. The same architecture and decoding procedure handle both input modalities.
At inference, generated tokens are de-tokenized into a valid SVG path and rasterized for preview.

\begin{figure*}
    \centering
    \includegraphics[width=1.0\linewidth]{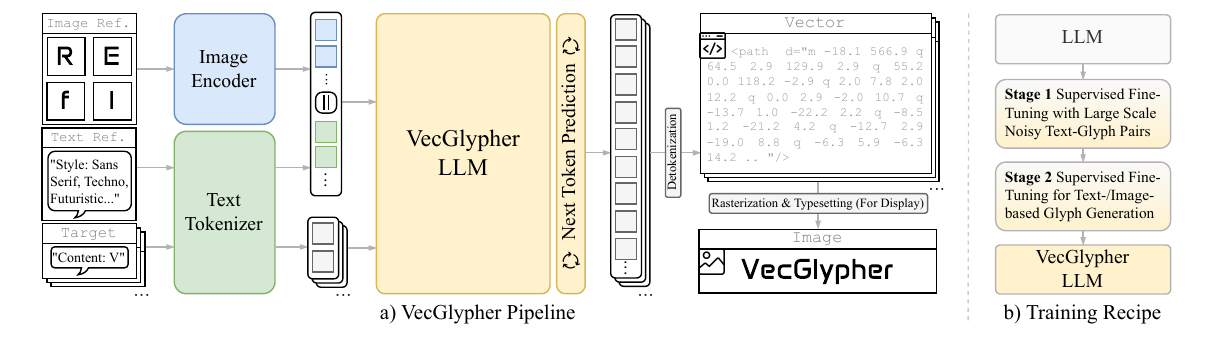}
    \caption{
    \textbf{VecGlypher pipeline and training recipe.} a) A text tokenizer or image encoder condition the LLM ($||$ denotes mutually exclusive choice), which predicts the next SVG token until the path is produced; detokenization yields SVG paths that we rasterize for display only. b) Training is two-stage: 1) SFT on Envato (text-referenced only) teaches SVG syntax and long-horizon geometry; 2) SFT on Google Fonts (text- or image-referenced) aligns geometry to appearance instructions. No raster denoisers or post-optimizers are used. 
    }
    \label{fig:method}
\end{figure*}

Achieving reliable glyph drawing with an LLM requires both the right training recipe and typography-aware data engineering. We curate complementary corpora and adopt a two-stage scheme (\cref{fig:method}): \textbf{Stage 1} performs large-scale continuation on noisy text-glyph pairs from a 39K-font Envato collection~\cite{envatoelementsfonts} to \emph{teach the model to draw}---establishing robust SVG syntax, long-horizon coordinate prediction, and character-conditioned geometry. \textbf{Stage 2} post-trains on a smaller but higher-quality 2.5K-font Google Fonts set~\cite{googlefonts} with expert descriptors and image exemplars, sharpening the mapping from textual concepts or image exemplers to glyph geometry and enabling \emph{unified} generation for both modalities.

To support stable long-sequence decoding, we design a typography-specific preprocessing pipeline: de-duplicate near-identical fonts; remove malformed or excessively long paths; normalize all glyphs to a consistent coordinate system; serialize each glyph to a canonical SVG \texttt{<path>} representation; and tokenize commands and coordinates with fixed-precision quantization. These steps lower sequence complexity and reduce error propagation during decoding, turning a general-purpose multimodal LLM into a competent glyph drawer.

\Cref{fig:teaser} summarizes the contrast with prior paradigms and our resulting pipeline. Compared with dual encoder-decoder systems tailored to image-referenced generation or cascaded image-then-vector diffusion approaches, \textbf{VecGlypher} collapses glyph synthesis into a single autoregressive program that emits vector tokens directly. This yields three practical advantages: (1) no exemplar sheet is required---text alone suffices---while still supporting exemplars when available; (2) avoiding raster intermediates eliminates vectorization artifacts by generating valid SVG in one pass; and (3) the language-modeling formulation scales naturally---larger backbones, more continuation data, and improved tokenization translate into stronger glyph geometry and generalization.

We evaluate these claims extensively. Both qualitatively and quantitatively, \textbf{VecGlypher} produces cleaner, more stylistically faithful glyphs than specialized \textit{image-referenced} baselines, and succeeds on \textit{text-referenced} tasks where state-of-the-art LLMs and vector-graphics LLMs fail to draw at all. Ablations show that (i) model scale is a primary driver of vector fidelity and style consistency, and (ii) large-scale continuation in Stage~1 delivers tangible out-of-distribution gains beyond what limited expert-annotated data alone can offer.

Our contributions are summarized as follows:

$\bullet$  \textbf{Unified formulation.} We recast vector glyph generation as language modeling over SVG path tokens and present \textbf{VecGlypher}, a single multimodal LLM that supports both text- and image-referenced conditioning while directly emitting vector code.

$\bullet$ \textbf{Two-stage training recipe.} We curate two complementary datasets (39K noisy fonts; 2.5K expert-annotated fonts) and show that large-scale continuation followed by targeted post-training is key to drawing high-quality glyphs from either modality.

$\bullet$ \textbf{Empirical validation.} Through comprehensive evaluations, we demonstrate substantial improvements over specialized image-referenced methods and show that general SVG-generation LLMs are insufficient for typography, whereas our approach follows textual descriptions and exemplar images with high fidelity.

Together, these results indicate that language models, properly trained and paired with typography-aware data engineering, can serve as a unified engine for vector glyph generation. \textbf{VecGlypher} lowers the barrier to font creation, lets users design with words or images, and provides a scalable foundation for future multimodal design tools.

\definecolor{lightgraytext}{gray}{0.75}  

\begin{table*}[t]
\centering
\begin{minipage}[t]{0.32\textwidth}
\centering
\caption{\textbf{Font filtering.}}
\label{tab:font-filter}
\scalebox{0.95}{
\tablestyle{2pt}{1.1}
\begin{tabular}{lcc}
\toprule
Filtering & {Google Fonts} & {Envato Fonts} \\
\midrule
\color{lightgraytext} Total fonts & \color{lightgraytext} 3,403 & \color{lightgraytext} 82,343 \\
Invalid / Lengthy & 2,989 & 70,248 \\
Duplicated  & 2,645 & 65,216 \\
MLLM OCR & 2,497 & 39,497 \\
\bottomrule
\end{tabular}%
}
\end{minipage}%
\hfill
\begin{minipage}[t]{0.32\textwidth}
\centering
\caption{\textbf{Font-level dataset statistics.}}
\label{tab:font-train-test}
\scalebox{0.95}{
\tablestyle{2pt}{1.0}
\begin{tabular}{lcccc}
\toprule
\multirow{2}{*}{Split} & \multicolumn{2}{c}{Google Fonts} & \multicolumn{2}{c}{Envato Fonts} \\
\cmidrule(lr){2-3}\cmidrule(lr){4-5}
 & {Font Family} & {Font} & {Font Family} & {Font} \\
\midrule
\color{lightgraytext} Total & \color{lightgraytext} 1,117 & \color{lightgraytext} 2,497 & \color{lightgraytext} 23,543 & \color{lightgraytext} 39,497 \\
Train & 997 & 2,243 & 22,543 & 37,926 \\
Test & 120 & 254 & 1,000 & 1,571 \\
\bottomrule
\end{tabular}%
}
\end{minipage}%
\hfill
\begin{minipage}[t]{0.32\textwidth}
\centering
\caption{\textbf{Glyph-level dataset statistics.}
}
\label{tab:glyph-train-test}
\scalebox{0.95}{
\tablestyle{2pt}{1.0}
\begin{tabular}{lcccc}
\toprule
\multirow{2}{*}{Split} & \multicolumn{2}{c}{Google Fonts} & \multicolumn{2}{c}{Envato Fonts} \\
\cmidrule(lr){2-3}\cmidrule(lr){4-5}
 & {Original} & {Filtering} & {Original} & {Filtering} \\
\midrule
\color{lightgraytext} Total & \color{lightgraytext} 159,808 & \color{lightgraytext} 157,899 & \color{lightgraytext} 2,527,718 & \color{lightgraytext} 2,495,363 \\
Train & 143,552 & 142,148 & 2,427,174 & 2,394,819 \\
Test & 16,256 & 15,751 & 100,544 & 100,544 \\
\bottomrule
\end{tabular}%
}
\end{minipage}

\vspace{1em} 
    \centering
    \includegraphics[width=1.0\linewidth]{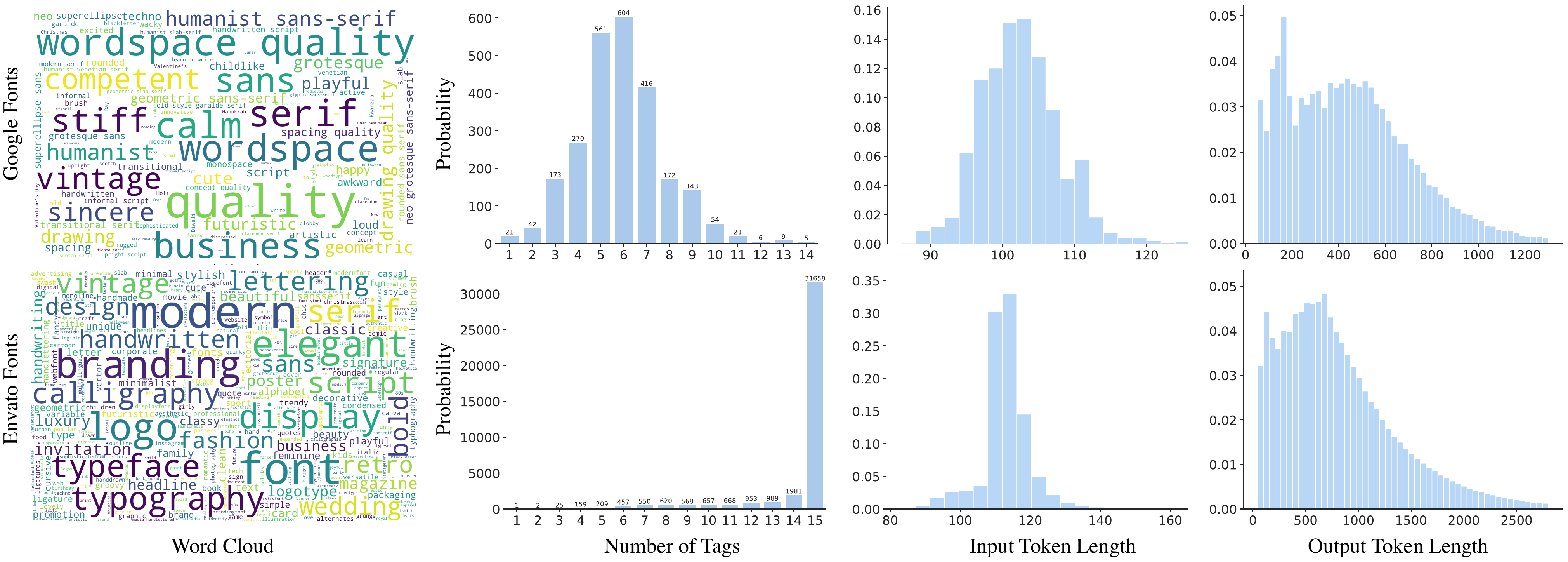}
    \caption{
    \textbf{Dataset statistics.} Word clouds (left) visualize tag vocabularies for \textit{Google Fonts} (expert-curated, appearance-focused) versus \textit{Envato} (noisier, marketing-oriented). The middle plots show the per-font number of tags (Google: concise; Envato: capped at $<$15). Right: distributions of input token length (tags) and output token length (SVG path tokens), with Envato exhibiting heavier-tailed outputs. These differences motivate our two-stage recipe: large-scale continuation on Envato, then instruction alignment on Google Fonts.
    }
    \label{fig:dataset_stats}          
\end{table*}

\section{Related Works}

\paragraph{Image-referenced glyph and font generation.}
Early systems synthesize glyph bitmaps and transfer style across characters from a few exemplar images~\cite{wang2020attribute2font,reddy2021multiimplicit,li2024hfhfont,yang2023fontdiffuser,he2024difffont,hayashi2019glyphgan}. While visually compelling, raster pipelines remain decoupled from vectorization, limiting editability and resolution-independent reuse. Vector-native methods instead predict parametric curves directly. \cite{lopes2019learned} sparked interest in structured, editable vector outputs; subsequent work predicts vector primitives from reference images to reconstruct style-consistent glyphs~\cite{wang2021deepvecfont}, improves contour fidelity and content-style disentanglement~\cite{wang2023deepvecfontv2}, jointly models curve geometry and filled regions~\cite{liu2023dualvector}, represents glyphs with signed distance fields~\cite{xia2023vecfontsdf}, and couples diffusion priors with a vector decoder for sparse-exemplar generation~\cite{thamizharasan2024vecfusion}. These specialized architectures are typically trained on moderate-scale corpora (e.g., \cite{wang2021deepvecfont,wang2023deepvecfontv2} $\approx$8K fonts; \cite{thamizharasan2024vecfusion} $\approx$1.4K). VecGlypher differs by combining substantially larger supervision (39K+2.5K fonts) with a language-model backbone to unify text or image exemplar-conditioned synthesis and vector decoding, yielding stronger cross-script generalization and directly editable glyph outputs.

\paragraph{Text-referenced glyph and font generation.}
A complementary line conditions font synthesis on language. Early datasets enabled modeling from style tags and textual attributes~\cite{chen2019large,odonovan2014exploratory}. Systems generate fonts from user-specified “impressions”~\cite{matsuda2021impressions2font,kang2022shared} or learn under incomplete tags~\cite{matsuda2022fontgeneration}. More recent work moves toward natural language, accepting free-form descriptions~\cite{wang2024typeface}, guiding diffusion with text descriptors~\cite{kang2024grifdm}, and supporting multimodal, interactive optimization for iterative design~\cite{tatsukawa2025fontcraft}. Most of these operate in the raster domain or emphasize retrieval or editing rather than producing structured, vector-native fonts with family-level consistency. In contrast, VecGlypher supports \emph{both} text and exemplar conditioning in a single model and emits vector glyphs directly, linking language-level intent to geometry without a separate vectorization stage.

\paragraph{Language models for vector graphics.}
Large language models demonstrate strong compositional reasoning, code generation, and multimodal instruction following~\cite{radford2018improving_gpt1,radford2019language_gpt2,brown2020language_gpt3,achiam2023gpt4,comanici2025gemini,anthropic2024claude,jaech2024openai_o1}. Text-to-vector approaches optimize or generate SVG-like structures from prompts, often via differentiable rendering and discrete primitives~\cite{jain2022vectorfusion,li2020diffvg,xu2022live,xing2023svgdreamer,wu2024chat2svg}. Vector-graphics LLMs go further by writing or editing SVG programs for icons and simple illustrations~\cite{rodriguez2023starvector,wu2023iconshop,xing2025empowering_llm4svg,yang2025omnisvg}. Despite promising results, these efforts generally target generic shapes, rely on small curated corpora, and do not address family-level consistency, Unicode coverage, or the typographic constraints unique to fonts. To our knowledge, VecGlypher is the \textit{first} unified LLM-based framework tailored to glyphs: it adapts an multimodal LLM to directly produce high-fidelity, editable vector glyphs and supports both text and image references within one model. At the 39K+2.5K data scale, this formulation closes the gap between language-conditioned intent and professional, vector-native type design.

\section{Method}

VecGlypher learns to autoregressively emit the SVG \verb|<path>| string of a target glyph from either a textual style description or a set of reference glyph images. Prior pipelines (i) translate images to vectors with separate encoder--decoders and a geometry post-optimizer~\cite{wang2021deepvecfont,wang2023deepvecfontv2,liu2023dualvector}, or (ii) cascade image and vector diffusions~\cite{thamizharasan2024vecfusion}. In contrast, VecGlypher \emph{unifies} style understanding and outline drawing within a single LLM conditioned on text or images. This removes intermediate mismatches and yields a \textit{single training objective}: \textbf{next-token prediction on SVG}.

\subsection{Preliminaries}

A \emph{glyph} is the vector outline of one character. We generate one glyph at a time as a single SVG \verb|<path>|. SVG represents a path as a sequence of drawing commands with arguments; uppercase commands use absolute coordinates and lowercase commands use relative coordinates. \cref{tab:svg_draw_command} summarizes the commands we allow; no other SVG attributes (e.g., fill, stroke) are used. During training and inference, each glyph is serialized as a \emph{string} of SVG drawing commands with \emph{one-decimal} precision.

\subsection{Data curation}
\label{sec3:method:data}


\paragraph{Envato fonts.} A large collection with up to fifteen \emph{noisy retrieval tags} per font (often names, authors, or marketing phrases). We use this corpus to \emph{teach} the model SVG syntax and geometry for glyph drawing.

\paragraph{Google Fonts.} A smaller but \emph{expert-tagged} set with appearance descriptors (e.g., humanist sans-serif, monoline, x-height). We use it for \emph{instruction following}, i.e., learning to follow style tags and to imitate styles from image references.

\paragraph{Dataset statistics.} According to \cref{fig:dataset_stats}, Google Fonts provides concise, informative tags and moderate output lengths; Envato exhibits a hard cap on tag counts and longer, heavy-tailed output sequences.

\subsection{Dataset processing}

\paragraph{Font processing.} We apply four filters to increase the quality of the training data: 

\begin{enumerate}
    \item \textbf{Character coverage:} remove fonts lacking alphanumerics (``0--9, a--z, A--Z'') such as symbols or pictograms. 
    \item \textbf{Length by pangram:} render the pangram ``\textit{The quick brown fox jumps over the lazy dog}'' as SVG; discard fonts whose total path length exceeds the 0.9 quantile. 
    \item \textbf{Deduplication:} drop fonts with identical pangram renderings. 
    \item \textbf{Multimodal LLM sanity check:} render ``GgAa'' with a state-of-the-art open-weight multimodal LLM (Qwen3-VL-30A3B-Instruct)~\cite{yang2025qwen3} and remove unicase fonts (small caps or oversized lowercase) and failed or ambiguous renders. \cref{tab:font-filter} reports resulting sizes.
\end{enumerate}
We split \emph{by font family} to measure cross-family generalization. In training we exclude extremely long glyphs ($>$0.9 quantile of tokenized path length). In testing we remove glyphs whose outlines duplicate those in training. \cref{tab:font-train-test} summarizes the splits.

\paragraph{Glyph processing.} We extract each glyph as an SVG \verb|<path>| and standardize as follows: 

\begin{enumerate}
    \item \textbf{Representation:} keep both absolute and relative commands; never elide consecutive command letters. 
    \item \textbf{Normalization:} rescale to \mbox{UPM=1000} and vertically align to a shared baseline.
    \item \textbf{Parsing:} retain only the \verb|d=""| attribute of \verb|<path>|; drop all other attributes (e.g., ``fill'', ``stroke''). 
    \item \textbf{Quantization:} round all coordinates to one decimal. 
\end{enumerate}

\noindent \cref{tab:glyph-train-test} gives glyph-level counts.

\subsection{Problem formulation}
\label{sec3:method:train_infer}

\paragraph{Training and inference.}
Let $x^{\text{t}}$ be the tokenized style description, $x^{\text{i}}=(x^{\text{i}}_1,\ldots,x^{\text{i}}_N)$ a set of $N$ reference glyph images, and $x^{\text{c}}$ the target character in ``0--9a--zA--Z''. The glyph is an SVG token sequence $y=(y_1,\ldots,y_T)$ whose concatenation forms a valid \verb|d| string. VecGlypher (parameters $\theta$) models
\begin{align*}
p_\theta\!\left(y \mid (x^{\text{t}} \parallel x^{\text{i}}), x^{\text{c}}\right)
= \prod_{t=1}^{T} p_\theta\!\left(y_t \;\middle|\; y_{<t}, (x^{\text{t}} \parallel x^{\text{i}}), x^{\text{c}}\right),
\end{align*}
where $(x^{\text{t}} \parallel x^{\text{i}})$ denotes text- or image-referenced conditioning. We minimize next-token cross-entropy on the \emph{SVG path text}. Consistent with~\cite{rodriguez2023starvector}, specialized SVG tokenization~\cite{xing2025empowering_llm4svg,yang2025omnisvg} adds complexity without measurable gains. At inference we use conservative decoding to favor syntactically valid SVG.

\paragraph{Output space.}
The model outputs the exact text of a single SVG \verb|<path>|. Because subword tokenization naturally covers command letters, separators, and numeric substrings, we simply detokenize to obtain the final \verb|d| string and render if needed (\cref{fig:method}a). We do \emph{not} apply raster denoisers, vector post-optimizers, or outline simplifiers~\cite{wang2021deepvecfont,wang2023deepvecfontv2,liu2023dualvector}.

\paragraph{Prompts.}
A strict system prompt enforces output structure and removes non-SVG tokens. For text-referenced generation, the input is a bag of style tags plus the target character; we randomly permute tag order during training to reduce positional bias. For image-referenced generation, we provide \mbox{1--8} reference glyph images from the same font, rendered at $192{\times}192$ with centering and uniform padding. The model transfers style and metrics from references to the requested content.

\subsection{Training recipe}

Our training uses two-stage supervised fine-tuning (\cref{fig:method}b).

\textbf{Stage 1: learning to draw (Envato).}
We perform supervised fine-tuning (SFT) on Envato~\cite{envatoelementsfonts} using \emph{text-referenced} samples only. Despite noisy tags, the scale and diversity force the model to master SVG glyph syntax and long-horizon geometry. Randomly sampled glyph-tag pairs are used as inputs; targets are gold \verb|<path>| strings. This stage markedly improves closure, winding consistency, and control-point placement.

\textbf{Stage 2: instruction following (Google Fonts).}
We then SFT on Google Fonts~\cite{googlefonts} with a \emph{mixture} of text- and image-referenced samples to align geometry with appearance instructions. Empirically, Stage~1 is essential: models initialized with it generalize better to unseen families and produce more stable outlines.

\section{Experiments}

\subsection{Implementation Details}
We fine-tune with AdamW at a base learning rate (LR) of $1\times10^{-5}$, weight decay of $0.01$, and scale the LR by the square root of the GPU scaling factor. A cosine schedule with 1\% warm-up decays the LR to zero. Envato is trained for one epoch and Google Fonts for three. For the 4B setting we use 8$\times$A100 with a global batch of 32; for 27B and 70B we use 32$\times$A100 with a global batch of 128. Decoding uses greedy sampling to assess raw generation capability. Input and conditioning details are in~\cref{sec3:method:train_infer}.

\subsection{Datasets and Evaluation}

\paragraph{Corpora and splits.}
We evaluate on Google Fonts, where expert-curated tags closely match glyph appearance, and use Envato as a large but noisy pre-training source. To test out-of-distribution (OOD) generalization, Google Fonts is split by \emph{family} so that styles from the same family never appear in both train and test. Additional details are in~\cref{sec3:method:data}.

\paragraph{Evaluation protocol.}
Unless noted, we report results on Google Fonts \emph{test families} only (cross-family OOD). Metrics:
\begin{enumerate}
\item \textbf{Relative OCR Accuracy (R-ACC).} We run Qwen3-VL-30A3B-Instruct~\cite{yang2025qwen3} as the OCR engine, prompting abstention for unrecognizable glyphs~\cite{chang2025oneig_bench-llm_ocr,geng2025x_omni-llm_ocr}. Accuracy on generated glyphs is normalized by \textit{OCR accuracy on the ground truths} and scaled by 100; scores can exceed 100 due to OCR variability (higher is better).
\item \textbf{Chamfer Distance (CD)}~\cite{borgefors1986distance_chamfer,borgefors2002hierarchical_chamfer}. We normalize both glyphs to $[-1,1]$ (preserving aspect ratio), uniformly sample 200 points per glyph, and compute the symmetric Chamfer distance, scaled by 100 (lower is better).
\item \textbf{CLIP}~\cite{radford2021learning_clip}. Text–image similarity (ViT-B/32) between the prompt tags and a rasterized preview (higher is better).
\item \textbf{DINO}~\cite{oquab2023dinov2}. Cosine \emph{similarity} of DINOv2 ViT-B/14 features between the generated and ground-truth glyphs, scaled by 100 (higher is better).
\item \textbf{FID}~\cite{heusel2017gans_fid}. Fréchet Inception Distance between generated and target glyph distributions (lower is better).
\end{enumerate}


\subsection{Ablations}

\begin{wrapfigure}{r}{0.55\textwidth}
    \centering
  \includegraphics[width=1.0\linewidth]{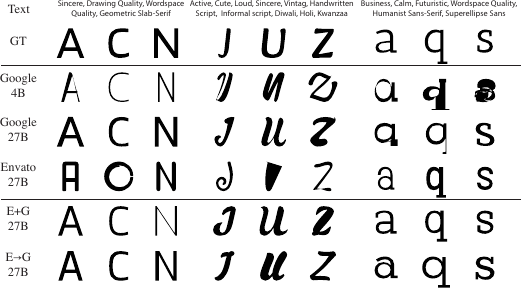}
  \caption{
    \textbf{Text-referenced ablations.} Representative text-to-glyph generations across model sizes and data regimens: ground truth (GT), Google-only 4B and 27B models, Envato-only 27B, mixed E+G 27B, and two-stage E$\to$G 27B. Using the same style tags across columns, scaling and the two-stage recipe yield cleaner closures, stable counters, and more faithful style. 
    Please refer to the supplementary materials for comprehensive results.
  }
    \label{fig:ablation:text}
\end{wrapfigure}
\paragraph{Text-referenced.}
We train Gemma3~\cite{team2025gemma} 4B and 27B on Google Fonts and study data sources at 27B; see \cref{tab:ablations:text_ref}.

\noindent$\bullet$ \textbf{Model size.} Scaling from 4B to 27B markedly improves geometry and fidelity: on Google-only training, R-ACC rises from 73.96/66.66 (relative/absolute coordinates) to 92.81/94.91, while CD drops from 4.29/3.75 to 2.31/1.98 and FID from 15.57/14.69 to 5.81/3.96. CLIP and DINO similarly improve.

\noindent$\bullet$ \textbf{Data source.} Using Envato only at 27B yields strong recognition (R-ACC $\approx$94) but weak geometry/fidelity, confirming that noisy supervision alone does not teach precise contours. Mixing Envato and Google in a single stage improves recognition yet remains inferior in geometry/fidelity to Google-only training. A two-stage recipe consistently performs best: with absolute coordinates we obtain R-ACC 101.0, CD 1.67, DINO 94.34, FID 3.47.

\noindent$\bullet$ \textbf{SVG representation.} At 27B, absolute coordinates slightly but consistently outperform relative across metrics. At 4B, relative coordinates can yield marginally higher R-ACC but worse geometry, indicating absolute coordinates become preferable as capacity increases.
We provide qualitative results in~\cref{fig:ablation:text}.

\paragraph{Image-referenced.}
We repeat the analysis for the image-referenced task; see \cref{tab:ablations:img_ref}.

\noindent$\bullet$ \textbf{Model size.} With Google-only training, 27B improves over 4B on all metrics (e.g., CD 1.41 vs.\ 2.11, FID 2.84 vs.\ 7.94 with absolute coordinates).

\noindent$\bullet$ \textbf{Data source.} Pre-training on \emph{text-referenced} Envato and then fine-tuning on \emph{image-referenced} Google improves geometry (CD 1.16--1.17) and fidelity (FID 2.51--2.55) versus Google-only. A joint second stage mixing text- and image-referenced Google is best: with absolute coordinates we reach R-ACC 99.12, CD 1.18, CLIP 26.07, DINO 95.82, and FID 2.32.

\begin{wrapfigure}{r}{0.55\textwidth}
    \centering
  \includegraphics[width=1.0\linewidth]{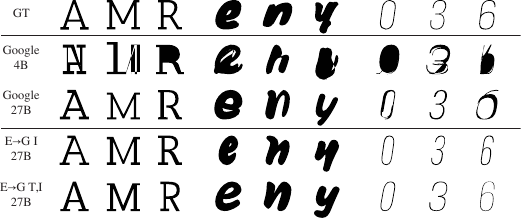}
  \caption{
\textbf{Image-referenced ablations.} Given 1--8 reference glyphs from a font, we compare Google-only 4B/27B with two-stage variants (E$\to$G I and E$\to$G T,I). The two-stage settings at 27B best transfer style and preserve thin structures and closures; Google-only baselines underperform in geometry. 
Please refer to the supplementary materials for comprehensive results.
  }
    \label{fig:ablation:image}
\end{wrapfigure}

\noindent$\bullet$ \textbf{SVG representation.} As above, absolute coordinates are equal or better for 27B; 4B sometimes gains slightly higher R-ACC with relative coordinates but with worse geometry.
The qualitative comparisons can be found in~\cref{fig:ablation:image}.

\begin{table}[t]
\centering

\caption{
\textbf{Text-referenced ablations on data and model size.} We compare relative (R) vs. absolute (A) coordinate serialization and several data regimens: Google-only (G), Envato-only (E), mixed E+G, and two-stage E$\to$G. 
Scaling and the two-stage E$\to$G training with absolute coordinates yield the best geometry and fidelity. 
}
\label{tab:ablations:text_ref}

\tablestyle{1pt}{1.0}
\begin{tabular}{y{30} x{15} x{20} x{35} x{25} x{30} x{30} x{30}}
\toprule[1.0pt] 
\textbf{Data} & \textbf{Size} & \textbf{Repr.} & \textbf{R-ACC$\,\uparrow$} & \textbf{CD$\,\downarrow$} & \textbf{CLIP$\,\uparrow$} & \textbf{DINO$\,\uparrow$} & \textbf{FID$\,\downarrow$} \\
\midrule
\multirow{4}{*}{Google}                  & \multirow{2}{*}{4B}   & R  & {\cellcolor[rgb]{0.914,0.965,0.941}}73.96 & 4.29                                     & 26.02                                     & {\cellcolor[rgb]{0.953,0.98,0.969}}90.27  & {\cellcolor[rgb]{0.796,0.918,0.859}}15.57  \\
                                         &    & A  & 66.66                                     & {\cellcolor[rgb]{0.902,0.961,0.933}}3.75 & {\cellcolor[rgb]{0.988,0.996,0.992}}26.04 & {\cellcolor[rgb]{0.988,0.996,0.992}}89.92 & {\cellcolor[rgb]{0.784,0.914,0.851}}14.69  \\
\cmidrule{2-8}
                                         & \multirow{2}{*}{27B}  & R  & {\cellcolor[rgb]{0.69,0.875,0.784}}92.81  & {\cellcolor[rgb]{0.616,0.843,0.733}}2.31 & {\cellcolor[rgb]{0.714,0.886,0.804}}26.38 & {\cellcolor[rgb]{0.682,0.871,0.78}}93.01  & {\cellcolor[rgb]{0.667,0.863,0.769}}5.81   \\
                                         &   & A  & {\cellcolor[rgb]{0.639,0.855,0.749}}94.91 & {\cellcolor[rgb]{0.471,0.784,0.631}}1.98 & {\cellcolor[rgb]{0.675,0.871,0.776}}26.43 & {\cellcolor[rgb]{0.584,0.831,0.71}}93.43  & {\cellcolor[rgb]{0.408,0.761,0.588}}3.96   \\
\midrule
\multirow{2}{*}{Envato}                  & \multirow{2}{*}{27B}  & R  & {\cellcolor[rgb]{0.675,0.871,0.776}}93.99 & {\cellcolor[rgb]{0.925,0.969,0.949}}3.89 & {\cellcolor[rgb]{0.667,0.867,0.769}}26.43 & 89.79                                     & 30.68                                      \\
                                         &   & A  & {\cellcolor[rgb]{0.678,0.871,0.776}}93.84 & {\cellcolor[rgb]{0.878,0.949,0.918}}3.63 & {\cellcolor[rgb]{0.62,0.847,0.737}}26.45  & {\cellcolor[rgb]{0.957,0.984,0.973}}90.24 & {\cellcolor[rgb]{0.863,0.941,0.906}}20.43  \\
\midrule
\multirow{2}{*}{E\;$+$\;G}    & \multirow{2}{*}{27B}  & R  & {\cellcolor[rgb]{0.663,0.863,0.769}}94.39 & {\cellcolor[rgb]{0.69,0.875,0.784}}2.56  & {\cellcolor[rgb]{0.706,0.882,0.8}}26.39   & {\cellcolor[rgb]{0.651,0.859,0.761}}93.17 & {\cellcolor[rgb]{0.671,0.867,0.773}}5.83   \\
                                         &   & A  & {\cellcolor[rgb]{0.604,0.843,0.725}}95.59 & {\cellcolor[rgb]{0.541,0.812,0.682}}2.14 & {\cellcolor[rgb]{0.604,0.843,0.725}}26.45 & {\cellcolor[rgb]{0.522,0.808,0.671}}93.66 & {\cellcolor[rgb]{0.533,0.812,0.675}}4.86   \\
\midrule
\multirow{2}{*}{E\,$\to$\,G} & \multirow{2}{*}{27B}  & R  & {\cellcolor[rgb]{0.396,0.757,0.58}}99.88  & {\cellcolor[rgb]{0.357,0.737,0.549}}1.71 & {\cellcolor[rgb]{0.38,0.749,0.569}}26.52  & {\cellcolor[rgb]{0.412,0.765,0.592}}94.07 & {\cellcolor[rgb]{0.353,0.737,0.549}}3.57   \\
                                         &   & A  & {\cellcolor[rgb]{0.341,0.733,0.541}}101.0 & {\cellcolor[rgb]{0.341,0.733,0.541}}1.67 & {\cellcolor[rgb]{0.341,0.733,0.541}}26.53 & {\cellcolor[rgb]{0.341,0.733,0.541}}94.34 & {\cellcolor[rgb]{0.341,0.733,0.541}}3.47   \\
\bottomrule[1.0pt]
\end{tabular}
\end{table}

\begin{table}[t]
\centering

\caption{
\textbf{Image-referenced ablations on data and model size.} With 1---8 reference glyphs, we study Google-only 4B/27B and two-stage variants that first SFT on Envato then fine-tune on Google (``E$\to$G I'' for image-only training; ``E$\to$G T,I'' for mixed training). At 27B, absolute coordinates achieve the strongest overall results, outperforming single-stage training. 
}
\label{tab:ablations:img_ref}

\tablestyle{1pt}{1.0}
 
\begin{tabular}{y{30} x{15} x{20} x{35} x{25} x{30} x{30} x{30}}
\toprule[1.0pt]
\textbf{Data} & \textbf{Size} & \textbf{Repr.} & \textbf{R-ACC$\,\uparrow$} & \textbf{CD$\,\downarrow$} & \textbf{CLIP$\,\uparrow$} & \textbf{DINO$\,\uparrow$} & \textbf{FID$\,\downarrow$} \\
\midrule
\multirow{4}{*}{Google}    & \multirow{2}{*}{4B}    & R  & {\cellcolor[rgb]{0.969,0.988,0.98}}83.48  & 2.36                                     & {\cellcolor[rgb]{0.988,0.996,0.992}}25.90 & {\cellcolor[rgb]{0.98,0.992,0.988}}93.26  & 9.38                                      \\
                           &     & A  & 81.94                                     & {\cellcolor[rgb]{0.922,0.965,0.945}}2.11 & 25.89                                     & 93.11                                     & {\cellcolor[rgb]{0.925,0.969,0.949}}7.94  \\
\cmidrule{2-8}
                           & \multirow{2}{*}{27B}   & R  & {\cellcolor[rgb]{0.733,0.894,0.816}}94.42 & {\cellcolor[rgb]{0.737,0.894,0.82}}1.53  & {\cellcolor[rgb]{0.722,0.886,0.808}}26.03 & {\cellcolor[rgb]{0.749,0.898,0.827}}94.88 & {\cellcolor[rgb]{0.737,0.894,0.816}}4.07  \\
                           &    & A  & {\cellcolor[rgb]{0.686,0.875,0.784}}96.46 & {\cellcolor[rgb]{0.698,0.878,0.792}}1.41 & {\cellcolor[rgb]{0.694,0.878,0.792}}26.04 & {\cellcolor[rgb]{0.71,0.882,0.8}}95.15    & {\cellcolor[rgb]{0.675,0.867,0.776}}2.84  \\
\midrule
\multirow{2}{*}{E$\to$G I}   & \multirow{2}{*}{27B}   & R  & {\cellcolor[rgb]{0.38,0.749,0.569}}98.90  & {\cellcolor[rgb]{0.361,0.741,0.553}}1.17 & {\cellcolor[rgb]{0.431,0.769,0.604}}26.06 & {\cellcolor[rgb]{0.455,0.78,0.624}}95.69  & {\cellcolor[rgb]{0.506,0.8,0.659}}2.51    \\
                           &    & A  & {\cellcolor[rgb]{0.553,0.82,0.69}}97.88   & {\cellcolor[rgb]{0.341,0.733,0.541}}1.16 & {\cellcolor[rgb]{0.498,0.796,0.651}}26.06 & {\cellcolor[rgb]{0.341,0.733,0.541}}95.84 & {\cellcolor[rgb]{0.541,0.812,0.682}}2.55  \\
\midrule
\multirow{2}{*}{E$\to$G T,I} & \multirow{2}{*}{27B}   & R  & {\cellcolor[rgb]{0.349,0.737,0.549}}99.08 & {\cellcolor[rgb]{0.451,0.776,0.616}}1.21 & {\cellcolor[rgb]{0.341,0.733,0.541}}26.07 & {\cellcolor[rgb]{0.486,0.792,0.643}}95.65 & {\cellcolor[rgb]{0.482,0.788,0.639}}2.48  \\
                           &    & A  & {\cellcolor[rgb]{0.341,0.733,0.541}}99.12 & {\cellcolor[rgb]{0.384,0.749,0.569}}1.18 & {\cellcolor[rgb]{0.404,0.761,0.584}}26.07 & {\cellcolor[rgb]{0.357,0.741,0.553}}95.82 & {\cellcolor[rgb]{0.341,0.733,0.541}}2.32  \\
\bottomrule[1.0pt]
\end{tabular}
\end{table}

\begin{table}[t]
\centering

\caption{
\textbf{Text-referenced comparisons with general LLMs.} We evaluate several proprietary/budget LLMs against VecGlypher on cross-family OOD fonts. Budget models rarely emit valid SVG; flagship models improve but still trail our approach. VecGlypher-70B (A) attains state-of-the-art performance. 
}
\label{tab:comp:text_ref}

\tablestyle{1pt}{1.0}

\begin{tabular}{y{75} x{31} x{30} x{30} x{30} x{30} x{30}}
\toprule[1.0pt]
\textbf{Model} &\textbf{R-ACC$\,\uparrow$} & \textbf{CD$\,\downarrow$} & \textbf{CLIP$\,\uparrow$} & \textbf{DINO$\,\uparrow$} & \textbf{FID$\,\downarrow$} \\
\midrule
GPT-5 Mini             & 4.17                                      & {\cellcolor[rgb]{0.871,0.949,0.91}}10.70 & 24.75                                     & {\cellcolor[rgb]{0.965,0.988,0.976}}79.65 & {\cellcolor[rgb]{0.878,0.949,0.918}}63.86  \\
Gemini-2.5 Flash       & {\cellcolor[rgb]{0.996,1,0.996}}4.89      & 13.78                                    & {\cellcolor[rgb]{0.863,0.945,0.906}}25.26 & 78.62                                     & 86.03                                      \\
Claude Haiku 4.5       & {\cellcolor[rgb]{0.776,0.91,0.847}}32.38  & {\cellcolor[rgb]{0.745,0.894,0.824}}7.60 & {\cellcolor[rgb]{0.769,0.906,0.839}}25.61 & {\cellcolor[rgb]{0.78,0.914,0.847}}84.69  & {\cellcolor[rgb]{0.725,0.886,0.812}}35.07  \\
\midrule
GPT-5                  & {\cellcolor[rgb]{0.682,0.875,0.78}}43.98  & {\cellcolor[rgb]{0.686,0.871,0.784}}6.12 & {\cellcolor[rgb]{0.678,0.871,0.776}}25.95 & {\cellcolor[rgb]{0.698,0.878,0.792}}86.92 & {\cellcolor[rgb]{0.694,0.875,0.788}}29.00  \\
Gemini 2.5 Pro         & {\cellcolor[rgb]{0.843,0.937,0.89}}24.04  & {\cellcolor[rgb]{0.773,0.906,0.843}}8.22 & {\cellcolor[rgb]{0.78,0.914,0.847}}25.57  & {\cellcolor[rgb]{0.812,0.925,0.871}}83.77 & {\cellcolor[rgb]{0.796,0.914,0.859}}47.82  \\
Claude Sonnet 4.5      & {\cellcolor[rgb]{0.663,0.867,0.769}}46.65 & {\cellcolor[rgb]{0.635,0.851,0.745}}5.28 & {\cellcolor[rgb]{0.663,0.863,0.769}}25.99 & {\cellcolor[rgb]{0.639,0.855,0.749}}88.31 & {\cellcolor[rgb]{0.596,0.835,0.718}}19.59  \\
\midrule
VecGlypher 70B T,R & {\cellcolor[rgb]{0.345,0.737,0.545}}100.1 & {\cellcolor[rgb]{0.341,0.733,0.541}}1.70 & {\cellcolor[rgb]{0.349,0.737,0.545}}26.53 & {\cellcolor[rgb]{0.353,0.737,0.549}}94.10 & {\cellcolor[rgb]{0.341,0.733,0.541}}3.45   \\
VecGlypher 70B T,A & {\cellcolor[rgb]{0.345,0.737,0.545}}100.4 & {\cellcolor[rgb]{0.341,0.733,0.541}}1.68 & {\cellcolor[rgb]{0.341,0.733,0.541}}26.54 & {\cellcolor[rgb]{0.341,0.733,0.541}}94.28 & {\cellcolor[rgb]{0.341,0.733,0.541}}3.34   \\
\midrule
VecGlypher 27B T,I,R & {\cellcolor[rgb]{0.349,0.737,0.545}}99.62 & {\cellcolor[rgb]{0.345,0.733,0.545}}1.77 & {\cellcolor[rgb]{0.349,0.737,0.549}}26.53 & {\cellcolor[rgb]{0.357,0.741,0.553}}93.97 & {\cellcolor[rgb]{0.341,0.733,0.541}}3.50   \\
VecGlypher 27B T,I,A & {\cellcolor[rgb]{0.341,0.733,0.541}}100.5 & {\cellcolor[rgb]{0.341,0.733,0.541}}1.72 & {\cellcolor[rgb]{0.349,0.737,0.549}}26.53 & {\cellcolor[rgb]{0.345,0.737,0.545}}94.22 & {\cellcolor[rgb]{0.341,0.733,0.541}}3.46   \\
\bottomrule[1.0pt]
\end{tabular}
\end{table}

\begin{table}[t]
\centering

\caption{
\textbf{Image-referenced comparisons with vector-font baselines.} Against DeepVecFont-v2~\cite{wang2023deepvecfontv2} and DualVector~\cite{liu2023dualvector}, VecGlypher-27B (T,I) delivers substantially higher recognizability and drastically lower geometric and distributional errors, demonstrating strong generalization to unseen styles. 
}
\label{tab:comp:img_ref}

\tablestyle{1pt}{1.0}

\begin{tabular}{y{75} x{31} x{30} x{30} x{30} x{30} x{30}}
\toprule[1.0pt]
\textbf{Model} &\textbf{R-ACC$\,\uparrow$} & \textbf{CD$\,\downarrow$} & \textbf{CLIP$\,\uparrow$} & \textbf{DINO$\,\uparrow$} & \textbf{FID$\,\downarrow$} \\
\midrule
DeepVecFont-v2         & 37.86                                     & {\cellcolor[rgb]{0.925,0.969,0.949}}14.58 & 24.81                                     & 79.41                                     & 115.5                                      \\
DualVector             & {\cellcolor[rgb]{0.898,0.961,0.929}}49.20 & 16.45                                     & {\cellcolor[rgb]{0.89,0.957,0.925}}25.07  & {\cellcolor[rgb]{0.996,1,0.996}}79.57     & {\cellcolor[rgb]{0.945,0.976,0.961}}105.5  \\
\midrule
VecGlypher 27B T,I,R & {\cellcolor[rgb]{0.345,0.737,0.545}}99.08 & {\cellcolor[rgb]{0.341,0.733,0.541}}1.21  & {\cellcolor[rgb]{0.341,0.733,0.541}}26.07 & {\cellcolor[rgb]{0.349,0.737,0.549}}95.65 & {\cellcolor[rgb]{0.341,0.733,0.541}}2.48    \\
VecGlypher 27B T,I,A & {\cellcolor[rgb]{0.341,0.733,0.541}}99.12 & {\cellcolor[rgb]{0.341,0.733,0.541}}1.18  & {\cellcolor[rgb]{0.345,0.737,0.545}}26.07 & {\cellcolor[rgb]{0.341,0.733,0.541}}95.82 & {\cellcolor[rgb]{0.341,0.733,0.541}}2.32    \\
\bottomrule[1.0pt]
\end{tabular}
\end{table}

\subsection{Comparisons to Baselines}

\begin{figure*}[t]
  \centering
  \includegraphics[width=1.0\linewidth]{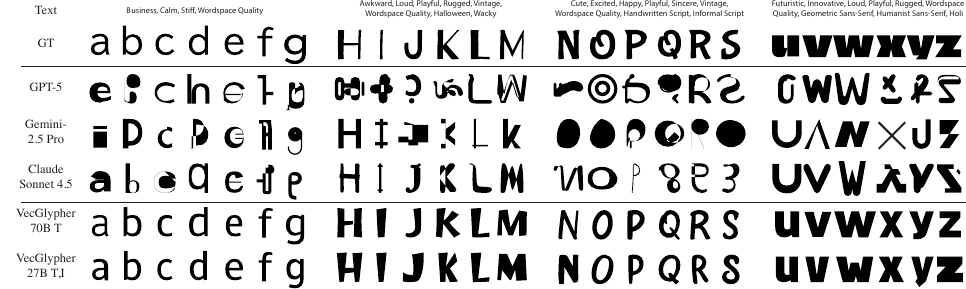}
  \caption{
  \textbf{Text-referenced comparisons to general LLMs.} Rows show GT, three flagship/budget multimodal LLMs, and our VecGlypher models. General LLMs often fail to output valid, closed paths or the requested character, whereas VecGlypher (27B/70B) consistently produces watertight, stylistically faithful vectors for the same prompts. 
  More results can be found in the supplementary materials.
  }
  \label{fig:comp:text}
\end{figure*}

\paragraph{Text-referenced.}
We compare to general-purpose proprietary LLMs and our VecGlypher models (\cref{tab:comp:text_ref}). Budget-tier models (e.g., GPT-5 Mini, Gemini-2.5 Flash) rarely output valid SVGs. Flagship models perform better but still struggle to follow the requested character and to close contours. VecGlypher substantially outperforms all baselines: VecGlypher-27B (Gemma3-27B) attains R-ACC 100.5, CD 1.72, DINO 94.22, and FID 3.46; VecGlypher-70B (Llama3.3-70B~\cite{grattafiori2024llama}) further improves CD/FID to 1.68/3.34 with R-ACC 100.4. Relative to the strongest baseline (Claude Sonnet 4.5), VecGlypher-70B delivers $\sim$2.15$\times$ higher R-ACC, 68\% lower CD, and 83\% lower FID, while improving DINO by 6 points. These results highlight the current limitations of general LLMs for glyph generation and the effectiveness of our unified, two-stage training.
\cref{fig:comp:text} illustrates the qualitative comparison against the baselines.

\begin{figure*}[t]
  \centering
  \includegraphics[width=1.0\linewidth]{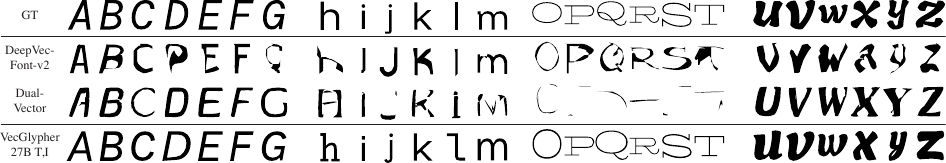}
  \caption{\textbf{Image-referenced comparisons to vector-font baselines.} On unseen font families, DeepVecFont-v2~\cite{wang2023deepvecfontv2} and DualVector~\cite{liu2023dualvector} struggle with thin strokes and contour closure. VecGlypher-27B (T,I) preserves delicate structures, follows target content, and closes contours.
  Please refer to the supplementary materials for the comprehensive comparisons.
  }
  \label{fig:comp:image}
\end{figure*}

\paragraph{Image-referenced.}
We compare with dedicated vector-glyph methods DeepVecFont-v2~\cite{wang2023deepvecfontv2} and DualVector~\cite{liu2023dualvector} (\cref{tab:comp:img_ref}). These approaches generalize poorly to unseen structures (e.g., thin strokes
). In contrast, VecGlypher-27B achieves R-ACC 99.12, CD 1.18, and FID 2.32, improving over the best baseline by $\sim$2$\times$ R-ACC, 92\% lower CD, and 97.8\% lower FID. Qualitative results show our outputs close contours, preserve thin structures, and follow target content faithfully.
\cref{fig:comp:image} illustrates the qualitative comparison against the baselines.

\paragraph{More qualitative results.} We include additional full qualitative samples across the full test set. Please refer to our supplementary material for thorough examination.

\section{Discussions and Conclusions}

\paragraph{Discussions.} VecGlypher exposes a sharp domain gap: LLMs that can write SVG icons rarely produce typographically valid glyphs. We posit this stems from training corpora that lack glyph programs, encouraging memorization of pictorial contours rather than learning vector-structural rules. Consequently, unified vector glyph generation is foremost a data problem: if next-generation LLMs ingest large, diverse corpora of vector glyph programs annotated with typographic constraints (closure, winding, counters, etc.), the capability should emerge with instruction tuning.

Scale still matters. Small models, even cost-efficient proprietary LLMs (see \cref{tab:comp:text_ref}), remain brittle; current evidence points to $\sim$30B parameters for stable quality. That operating point could drop with improved vector tokenization, constrained decoding, and lightweight geometric adapters once glyph-centric data is available. Progress also depends on measurement: composite proxy rewards that mix vector/raster geometry, topology, typographic regularity, and recognizability~\cite{tatsukawa2024fontclip} enable best-of-N sampling~\cite{wang2022self_best_of_n} and RL fine-tuning~\cite{guo2025deepseek_r1} from text or image cues.

Our current scope (``0--9, a--z, A--Z'') highlights the limits of closed-set vector methods~\cite{wang2021deepvecfont,wang2023deepvecfontv2,thamizharasan2024vecfusion,xia2023vecfontsdf}, which hinder the part sharing needed for diacritics, pictographs, and cursive writing. A path forward is to adopt compositional encodings
and trajectory-aware stroke programs. With such encodings and the right data, extending from Latin alphanumerics to open-ended writing systems appears to be a problem of scale rather than principle.

\paragraph{Conclusions.} We introduced VecGlypher, a unified multimodal language model that directly generates vector glyphs from text or exemplar images. Built on a dedicated data pipeline and a two-stage training recipe that combines large-scale noisy continuation with expert-tagged post-training, VecGlypher achieves substantially improved recognizability, geometry, and distributional fidelity than both general-purpose LLMs and specialized baselines in cross-family OOD evaluations. 
Ablation studies underline the importance of model scale, absolute-coordinate serialization, and staged supervision for stable SVG decoding and faithful style transfer. 
By lowering the barrier to font creation and coupling language, imagery, and vector geometry, VecGlypher positions LLMs as a scalable foundation for future multimodal design tools.

\clearpage
{
    \small
    \bibliographystyle{ieeenat_fullname}
    \bibliography{main}
}

\clearpage
\appendix
	\section*{Appendix}\label{sec:appendix}

\section{Dataset Statistics (Cont')}

\paragraph{Training data before filtering.}

Figure~1 visualizes the token-length distributions of the \emph{training} corpora \emph{before} applying the typography-specific filtering described in Sec.~3.3 of the main paper.

For each font, we compute:
\begin{itemize}
\item \textbf{Input token length}: the number of tokens in the serialized style description (for text-referenced samples).
\item \textbf{Output token length}: the number of tokens in the tokenized SVG \texttt{<path>} string for each glyph.
\end{itemize}

The top row of Fig.~1 shows the distributions for Google Fonts; the bottom row shows Envato.

\paragraph{Heavy-tailed SVG sequences.}
On both corpora, input style tokens are relatively concentrated around a narrow range (roughly one to two dozen tags per font, consistent with Fig.~2 in the main text).

In contrast, the \emph{SVG path} token lengths are strongly long-tailed, especially for Envato:
a small fraction of fonts contain glyphs whose serialized paths span tens of thousands of tokens.
These extremely long sequences mostly correspond to malformed outlines, redundant contour duplication, or decorative symbols.

Such heavy tails are problematic for autoregressive training: they increase sequence entropy, cause unstable gradients, and exacerbate error accumulation during decoding. This motivates the length-based pruning strategy (“Length by pangram”) described in Sec.~3.3 of the main paper.

\paragraph{Filtered test-set statistics.}

Fig.~2 reports analogous statistics for the \emph{testing} split, after all filtering steps and split-by-family partitioning.
Compared to Fig.~1, the output token distributions are significantly better behaved:

\begin{itemize}
\item The vast majority of glyphs fall into a moderate length regime, enabling stable training and evaluation.
\item Google Fonts remains more compact than Envato, consistent with its expert-curated nature and more regular outlines.
\end{itemize}

These plots confirm that the preprocessing pipeline successfully removes pathological fonts while retaining a broad diversity of typographic structure:
after filtering we keep 2{,}497 Google Fonts and 39{,}497 Envato fonts, with 157{,}899 and 2{,}495{,}363 glyphs respectively.
We exclude Envato fonts from testing because their tags are generally noisy and lack meaningful visual descriptions.


\begin{figure}[h]
    \centering
    \begin{subfigure}[t]{0.49\linewidth}
        \centering
        \includegraphics[width=\linewidth]{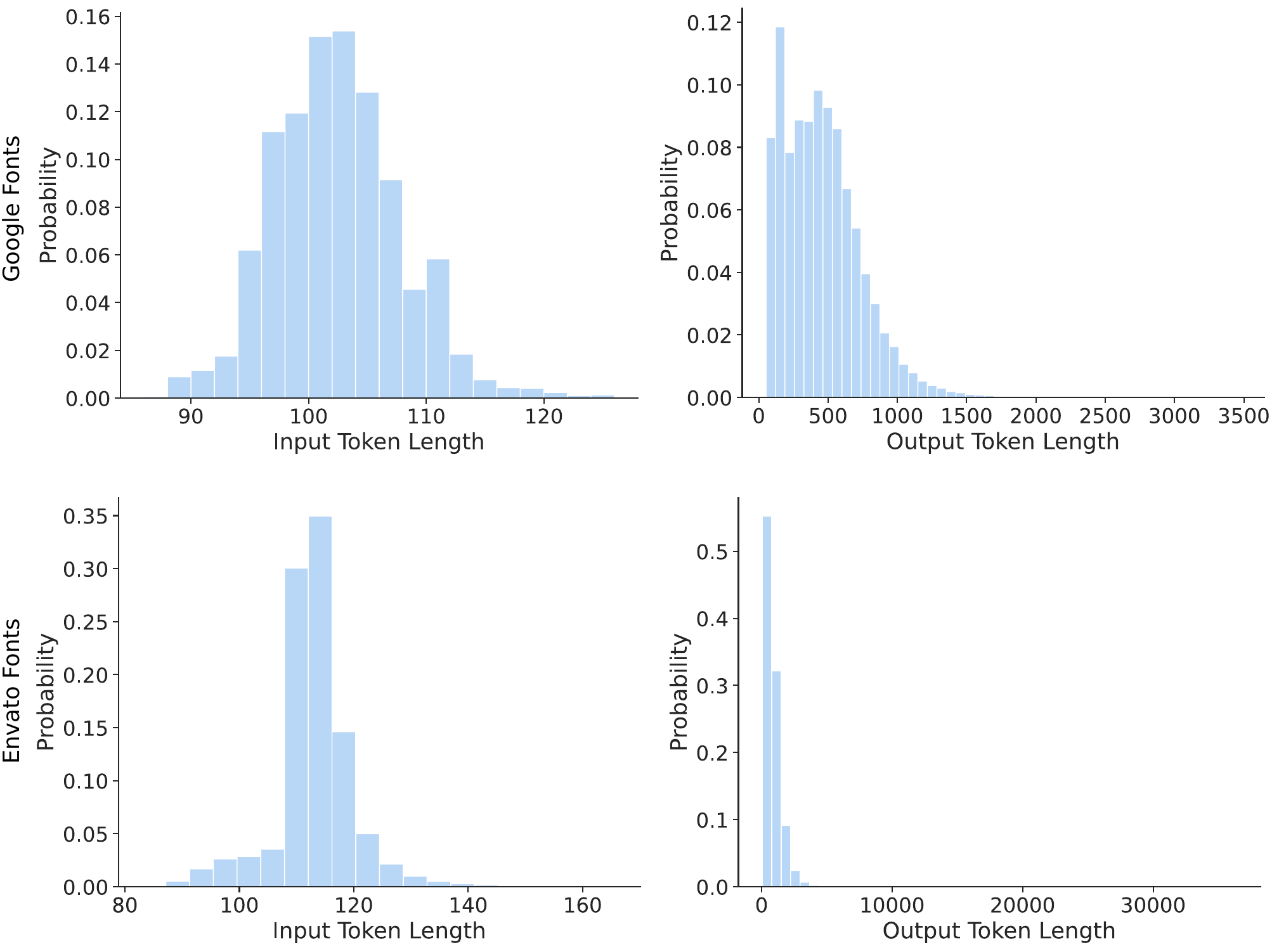}
        \caption{Dataset Statistics for training set before filtering.}
        \label{fig:dataset_stats-token_len-train-before_filter}
    \end{subfigure}
    \hfill
    \begin{subfigure}[t]{0.49\linewidth}
        \centering
        \includegraphics[width=\linewidth]{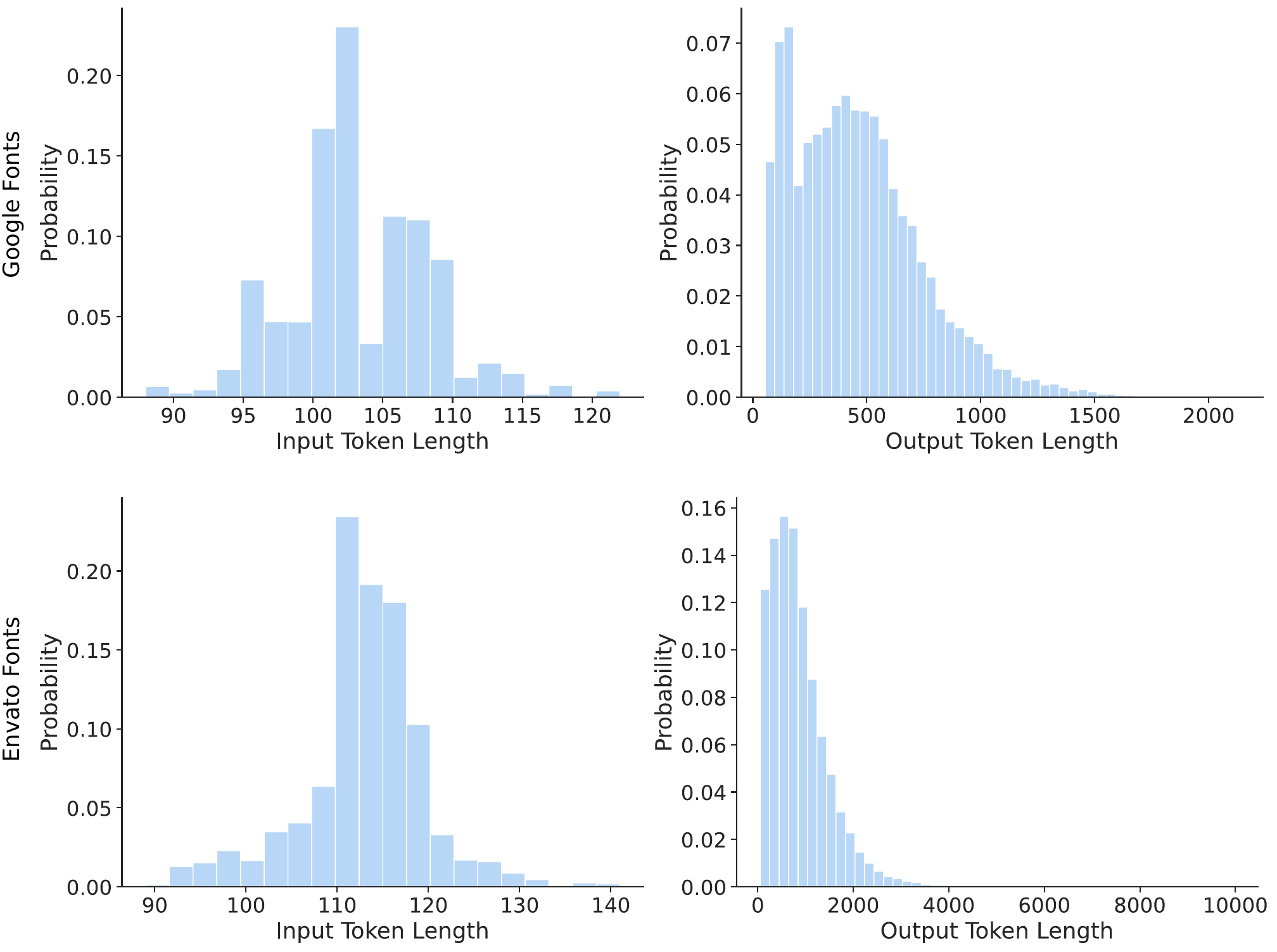}
        \caption{Dataset Statistics for testing set.}
        \label{fig:dataset_stats-token_len-test}
    \end{subfigure}
\end{figure}

\section{Prompt Templates and Samples}

We use a strict prompting scheme to encourage syntactically valid SVG paths and to cleanly separate role instructions, font style descriptions, and target content. The main paper briefly mentions a “strict system prompt” for SVG-only outputs;
here we describe the full templates, which are summarized in Table~1 of the supplementary material. The system prompt and the instruction prompts for both text- and image-referenced vector glyph generation. 
The ``\{\{FONT STYLE\}\}'' is the bag of style tags, 
and the ``\{\{GLYPH CHARACTER\}\}'' is a single given character in ``0–9, a–z, A–Z''.

\begin{table*}[h]

\centering
\begin{tcolorbox}[title=System Prompt, colback=white, colframe=black, width=\textwidth, breakable]
\label{tab:supp:prompts}
You are a specialized vector glyph designer creating SVG path elements.

Critical requirements:

- Each glyph must be a complete, self-contained $<$path$>$ element, in reading order of the given text.

- Terminate each $<$path$>$ element with a newline character.

- Output ONLY valid SVG $<$path$>$ elements.
\end{tcolorbox}
\vspace{+3mm}

\begin{tcolorbox}[title=Instruction Prompt for Text-Referenced Vector Glyph Generation, colback=white, colframe=black, width=\textwidth, breakable]
Font design requirements: \{\{FONT STYLE\}\}.

Text content: \{\{GLYPH CHARACTER\}\}.
\end{tcolorbox}
\vspace{+0mm}

\begin{tcolorbox}[title=Instruction Prompt for Image-Referenced Vector Glyph Generation, colback=white, colframe=black, width=\textwidth, breakable]
Font design requirements: faithfully match the provided reference images for style and metrics.

Text content: \{\{GLYPH CHARACTER\}\}.
\end{tcolorbox}
\vspace{+0mm}

\begin{tcolorbox}[title=Samples of instruction prompts for text-referenced vector glyph generation 1, colback=white, colframe=black, width=\textwidth, breakable]
Font design requirements: stiff, wordspace quality, superellipse sans, rounded sans-serif, 400 weight, sans-serif, normal style, display, neo grotesque sans-serif, drawing quality, cute.

Text content: V.
\end{tcolorbox}
\vspace{+0mm}

\begin{tcolorbox}[title=Samples of instruction prompts for text-referenced vector glyph generation 2, colback=white, colframe=black, width=\textwidth, breakable]
Font design requirements: wordspace quality, rounded sans-serif, sans-serif, competent, grotesque sans, 900 weight, italic style, calm.

Text content: b.
\end{tcolorbox}
\vspace{+0mm}

\begin{tcolorbox}[title=Samples of instruction prompts for text-referenced vector glyph generation 3, colback=white, colframe=black, width=\textwidth, breakable]
Font design requirements: fancy, 400 weight, vintage, sophisticated, handwritten script, brush, wordspace quality, handwriting, sincere, informal script, drawing quality, excited, normal style, cute, artistic.

Text content: 6.
\end{tcolorbox}

\end{table*}


\section{Additional Metrics}
The main paper evaluates VecGlypher and baselines with Relative OCR Accuracy (R-ACC), Chamfer Distance (CD), CLIP similarity, DINO similarity, and FID.

Here we introduce several additional metrics that capture complementary aspects of glyph quality and provide full results in Tables~3–6 of the supplementary.

All raster-based metrics are computed on 
192x192 grayscale renderings, using the same rasterization pipeline as for the qualitative figures.

We further introduce:

\begin{itemize}
\item \textbf{FID (C).}
In addition to FID computed in Inception space, we report \textbf{FID (C)}, where both real and generated glyphs are embedded with CLIP ViT-B/32 features before computing the Fréchet distance.
This emphasizes semantic similarity between the stylized glyph images and is more sensitive to high-level appearance than to low-level pixel noise.

\item \textbf{R-ACC (U)}: R-ACC in the main paper measures OCR accuracy normalized by the accuracy on ground-truth glyphs, allowing scores slightly above 100 because of OCR variability.

\item \textbf{R-ACC (U)}: The OCR outputs are case-normalized: upper- and lowercase characters that share the same identity (e.g., “a” vs “A”) are treated as correct. This disentangles shape-level recognition errors from case mismatches.

\item \textbf{CD (T) and CD (ST)}: Our primary Chamfer Distance (CD) computes a symmetric distance between two point clouds sampled from the SVG outlines after normalizing each glyph to the [$-1,1$] box.

\item \textbf{CD (T)}: we first align prediction to ground truth via Iterative Closest Point (ICP) optimizing only a 2D translation. This compensates for small global shifts, e.g., from imperfect baseline alignment.
\item \textbf{CD (ST)}: we run ICP optimizing both translation and isotropic scale. This is suitable for italic or script fonts where rotation is ambiguous but overall scale may drift.
We intentionally do \emph{not} optimize rotation in either variant, since many italic fonts are skewed and rotational alignment could distort their intended slant.

\item \textbf{{L2}}: {L2} is the mean squared error between rasterized predictions and ground truths, averaged over pixels and glyphs. It captures dense per-pixel discrepancies but is insensitive to human perception.

\item \textbf{{LPIPS}}: We report Learned Perceptual Image Patch Similarity (LPIPS) using a standard VGG backbone. LPIPS measures perceptual distance between images by comparing deep feature activations, correlating better with human judgement than L2.

\item \textbf{{PSNR and SSIM}}: We additionally report Peak Signal-to-Noise Ratio (PSNR) and Structural Similarity Index (SSIM), both computed on grayscale rasterizations. PSNR summarizes global reconstruction fidelity; SSIM emphasizes local luminance, contrast, and structural agreement.

\end{itemize}


\section{Additional Baselines}

To complement the proprietary LLMs evaluated in the main paper, we benchmark two classes of publicly available models on text-referenced glyph generation:

\begin{itemize}
\item \textbf{Open-weight multimodal LLMs}:
Llama~3.3 70B Instruct, Gemma~3 27B IT, and Qwen~3 30B A3B Instruct (2507).
These models are strong general-purpose multimodal assistants.
\item \textbf{Vector-graphics LLM}:
OmniSVG, an open-source model trained specifically for SVG generation and image vectorization. We skip LLM4SVG and StarVector since their text-to-SVG models are not publicly released.
\end{itemize}

\noindent Overall, we observe:

\begin{itemize}
\item \textbf{Extremely low recognizability.} R-ACC and R-ACC(U) remain close to zero for all open-weight LLMs and for OmniSVG, meaning that the OCR engine rarely recognizes the intended characters.
\item \textbf{Poor geometry and image quality.} CD, CD(T), and CD(ST) are an order of magnitude higher than those of VecGlypher, while L2, LPIPS, and SSIM indicate severe geometric distortions or failure to render the glyph at all.
\item \textbf{Limited benefit from SVG specialization.} OmniSVG, despite being trained for SVG icons and vectorization, produces particularly poor results for glyphs: a large fraction of outputs are invalid paths or generic shapes unrelated to the requested character or style.
\end{itemize}

These trends mirror the conclusions of Sec.~1 and Sec.~5 in the main paper: off-the-shelf LLMs and vector-graphics LLMs that perform well on icons or simple SVG drawings do not transfer to typography, which imposes stricter geometric, stylistic, and topological constraints.


\section{Additional Qualitative Results}

Please refer to the HTML pages for comprehensive results. They include both ablation studies and comparisons for text-referenced and image-referenced vector-glyph generation tasks.

\begin{table*}[h]
\centering
\caption{Text-referenced ablations on data and model size.}
\resizebox{\linewidth}{!}{%
\begin{tabular}{lll|ccccc|cccc|cccc} 
\toprule[1pt]
Data                                     & Size & Repr. & {R-ACC}                   & {CD}                     & {CLIP}                  & {DINO}                    & {FID}                   & {FID (C)}               & {R-ACC (U)}                & {CD (T)}                 & {CD (ST)}                & {L2}                    & {LPIPS}                 & {PSNR}                 & {SSIM}                   \\
\midrule
\multirow{4}{*}{Google}                  & 4B   & Rel.  & {\cellcolor[rgb]{0.914,0.965,0.941}}73.96 & 4.29                                     & 26.02                                     & {\cellcolor[rgb]{0.953,0.98,0.969}}90.27  & {\cellcolor[rgb]{0.796,0.918,0.859}}15.57 & {\cellcolor[rgb]{0.933,0.973,0.953}}1.84  & {\cellcolor[rgb]{0.918,0.969,0.941}}75.51  & 3.83                                     & 4.40                                     & {\cellcolor[rgb]{0.91,0.961,0.937}}0.196  & {\cellcolor[rgb]{0.882,0.953,0.918}}0.245 & {\cellcolor[rgb]{0.855,0.941,0.902}}8.30 & {\cellcolor[rgb]{0.886,0.953,0.922}}0.643  \\
                                         & 4B   & Abs.  & 66.66                                     & {\cellcolor[rgb]{0.902,0.961,0.933}}3.75 & {\cellcolor[rgb]{0.988,0.996,0.992}}26.04 & {\cellcolor[rgb]{0.988,0.996,0.992}}89.92 & {\cellcolor[rgb]{0.784,0.914,0.851}}14.69 & 2.19                                      & 68.68                                      & {\cellcolor[rgb]{0.91,0.961,0.937}}3.36  & {\cellcolor[rgb]{0.906,0.961,0.933}}3.85 & {\cellcolor[rgb]{0.965,0.984,0.973}}0.205 & {\cellcolor[rgb]{0.914,0.965,0.941}}0.251 & {\cellcolor[rgb]{0.882,0.953,0.922}}8.14 & {\cellcolor[rgb]{0.949,0.98,0.965}}0.631   \\
                                         & 27B  & Rel.  & {\cellcolor[rgb]{0.69,0.875,0.784}}92.81  & {\cellcolor[rgb]{0.616,0.843,0.733}}2.31 & {\cellcolor[rgb]{0.714,0.886,0.804}}26.38 & {\cellcolor[rgb]{0.682,0.871,0.78}}93.01  & {\cellcolor[rgb]{0.667,0.863,0.769}}5.81  & {\cellcolor[rgb]{0.659,0.863,0.765}}0.440 & {\cellcolor[rgb]{0.694,0.878,0.788}}93.40  & {\cellcolor[rgb]{0.612,0.839,0.729}}1.96 & {\cellcolor[rgb]{0.62,0.843,0.733}}2.32  & {\cellcolor[rgb]{0.686,0.871,0.784}}0.158 & {\cellcolor[rgb]{0.69,0.875,0.788}}0.212  & {\cellcolor[rgb]{0.698,0.878,0.792}}9.21 & {\cellcolor[rgb]{0.69,0.875,0.788}}0.679   \\
                                         & 27B  & Abs.  & {\cellcolor[rgb]{0.639,0.855,0.749}}94.91 & {\cellcolor[rgb]{0.471,0.784,0.631}}1.98 & {\cellcolor[rgb]{0.675,0.871,0.776}}26.43 & {\cellcolor[rgb]{0.584,0.831,0.71}}93.43  & {\cellcolor[rgb]{0.408,0.761,0.588}}3.96  & {\cellcolor[rgb]{0.459,0.78,0.624}}0.250  & {\cellcolor[rgb]{0.655,0.863,0.765}}95.30  & {\cellcolor[rgb]{0.478,0.788,0.639}}1.69 & {\cellcolor[rgb]{0.486,0.792,0.643}}2.00 & {\cellcolor[rgb]{0.608,0.839,0.729}}0.152 & {\cellcolor[rgb]{0.592,0.835,0.718}}0.204 & {\cellcolor[rgb]{0.58,0.831,0.71}}9.50   & {\cellcolor[rgb]{0.608,0.843,0.729}}0.686  \\
\midrule
\multirow{2}{*}{Envato}                  & 27B  & Rel.  & {\cellcolor[rgb]{0.675,0.871,0.776}}93.99 & {\cellcolor[rgb]{0.925,0.969,0.949}}3.89 & {\cellcolor[rgb]{0.667,0.867,0.769}}26.43 & 89.79                                     & 30.68                                     & {\cellcolor[rgb]{0.894,0.957,0.925}}1.63  & {\cellcolor[rgb]{0.675,0.871,0.776}}94.86  & {\cellcolor[rgb]{0.91,0.961,0.937}}3.35  & {\cellcolor[rgb]{0.914,0.965,0.941}}3.90 & 0.211                                     & 0.265                                     & 7.46                                     & 0.621                                      \\
                                         & 27B  & Abs.  & {\cellcolor[rgb]{0.678,0.871,0.776}}93.84 & {\cellcolor[rgb]{0.878,0.949,0.918}}3.63 & {\cellcolor[rgb]{0.62,0.847,0.737}}26.45  & {\cellcolor[rgb]{0.957,0.984,0.973}}90.24 & {\cellcolor[rgb]{0.863,0.941,0.906}}20.43 & {\cellcolor[rgb]{0.831,0.929,0.882}}1.31  & {\cellcolor[rgb]{0.663,0.867,0.769}}95.18  & {\cellcolor[rgb]{0.882,0.953,0.918}}3.22 & {\cellcolor[rgb]{0.871,0.945,0.91}}3.64  & {\cellcolor[rgb]{0.914,0.965,0.941}}0.197 & {\cellcolor[rgb]{0.933,0.973,0.953}}0.254 & {\cellcolor[rgb]{0.945,0.98,0.965}}7.78  & {\cellcolor[rgb]{0.922,0.969,0.945}}0.636  \\
\midrule
\multirow{2}{*}{$\text{E}+\text{G}$}   & 27B  & Rel.  & {\cellcolor[rgb]{0.663,0.863,0.769}}94.39 & {\cellcolor[rgb]{0.69,0.875,0.784}}2.56  & {\cellcolor[rgb]{0.706,0.882,0.8}}26.39   & {\cellcolor[rgb]{0.651,0.859,0.761}}93.17 & {\cellcolor[rgb]{0.671,0.867,0.773}}5.83  & {\cellcolor[rgb]{0.671,0.867,0.773}}0.458 & {\cellcolor[rgb]{0.678,0.871,0.776}}94.69  & {\cellcolor[rgb]{0.69,0.875,0.784}}2.20  & {\cellcolor[rgb]{0.69,0.875,0.784}}2.57  & {\cellcolor[rgb]{0.506,0.8,0.659}}0.147   & {\cellcolor[rgb]{0.525,0.808,0.671}}0.200 & {\cellcolor[rgb]{0.518,0.804,0.663}}9.60 & {\cellcolor[rgb]{0.494,0.796,0.647}}0.692  \\
                                         & 27B  & Abs.  & {\cellcolor[rgb]{0.604,0.843,0.725}}95.59 & {\cellcolor[rgb]{0.541,0.812,0.682}}2.14 & {\cellcolor[rgb]{0.604,0.843,0.725}}26.45 & {\cellcolor[rgb]{0.522,0.808,0.671}}93.66 & {\cellcolor[rgb]{0.533,0.812,0.675}}4.86  & {\cellcolor[rgb]{0.502,0.796,0.655}}0.289 & {\cellcolor[rgb]{0.6,0.839,0.725}}96.27    & {\cellcolor[rgb]{0.553,0.816,0.69}}1.84  & {\cellcolor[rgb]{0.541,0.812,0.682}}2.14 & {\cellcolor[rgb]{0.4,0.757,0.58}}0.141    & {\cellcolor[rgb]{0.412,0.761,0.588}}0.194 & {\cellcolor[rgb]{0.388,0.753,0.576}}9.81 & {\cellcolor[rgb]{0.384,0.753,0.573}}0.698  \\
\midrule
\multirow{2}{*}{$\text{E}\to\text{G}$} & 27B  & Rel.  & {\cellcolor[rgb]{0.396,0.757,0.58}}99.88  & {\cellcolor[rgb]{0.357,0.737,0.549}}1.71 & {\cellcolor[rgb]{0.38,0.749,0.569}}26.52  & {\cellcolor[rgb]{0.412,0.765,0.592}}94.07 & {\cellcolor[rgb]{0.353,0.737,0.549}}3.57  & {\cellcolor[rgb]{0.345,0.733,0.545}}0.142 & {\cellcolor[rgb]{0.392,0.753,0.576}}99.90  & {\cellcolor[rgb]{0.357,0.737,0.553}}1.44 & {\cellcolor[rgb]{0.361,0.741,0.557}}1.71 & {\cellcolor[rgb]{0.384,0.749,0.573}}0.141 & {\cellcolor[rgb]{0.412,0.761,0.588}}0.194 & {\cellcolor[rgb]{0.431,0.769,0.604}}9.74 & {\cellcolor[rgb]{0.4,0.757,0.584}}0.697    \\
                                         & 27B  & Abs.  & {\cellcolor[rgb]{0.341,0.733,0.541}}101.0 & {\cellcolor[rgb]{0.341,0.733,0.541}}1.67 & {\cellcolor[rgb]{0.341,0.733,0.541}}26.53 & {\cellcolor[rgb]{0.341,0.733,0.541}}94.34 & {\cellcolor[rgb]{0.341,0.733,0.541}}3.47  & {\cellcolor[rgb]{0.341,0.733,0.541}}0.135 & {\cellcolor[rgb]{0.341,0.733,0.541}}100.72 & {\cellcolor[rgb]{0.341,0.733,0.541}}1.40 & {\cellcolor[rgb]{0.341,0.733,0.541}}1.65 & {\cellcolor[rgb]{0.341,0.733,0.541}}0.138 & {\cellcolor[rgb]{0.341,0.733,0.541}}0.191 & {\cellcolor[rgb]{0.341,0.733,0.541}}9.88 & {\cellcolor[rgb]{0.341,0.733,0.541}}0.700  \\
\bottomrule[1pt]
\end{tabular}
}
\end{table*}

\begin{table*}[h]
\centering
\caption{Image-referenced ablations on data and model size.}
\resizebox{\linewidth}{!}{%
\begin{tabular}{lll|ccccc|cccc|cccc} 
\toprule[1pt]
Data                       & Size & Repr. & {R-ACC}                   & {CD}                     & {CLIP}                  & {DINO}                    & {FID}                  & {FID (C)}               & {R-ACC (U)}               & {CD (T)}                  & {CD (ST)}                & {L2}                    & {LPIPS}                 & {PSNR}                  & {SSIM}                   \\
\midrule
\multirow{4}{*}{Google}    & 4B   & Rel.  & {\cellcolor[rgb]{0.969,0.988,0.98}}83.48  & 2.36                                     & {\cellcolor[rgb]{0.988,0.996,0.992}}25.90 & {\cellcolor[rgb]{0.98,0.992,0.988}}93.26  & 9.38                                     & 1.08                                      & {\cellcolor[rgb]{0.976,0.992,0.984}}84.66 & 1.97                                      & 2.45                                     & {\cellcolor[rgb]{0.996,0.996,0.996}}0.155 & 0.206                                     & {\cellcolor[rgb]{0.996,1,1}}9.60          & {\cellcolor[rgb]{0.988,0.996,0.992}}0.684  \\
                           & 4B   & Abs.  & 81.94                                     & {\cellcolor[rgb]{0.922,0.965,0.945}}2.11 & 25.89                                     & 93.11                                     & {\cellcolor[rgb]{0.925,0.969,0.949}}7.94 & {\cellcolor[rgb]{0.996,0.996,0.996}}1.07  & 83.53                                     & {\cellcolor[rgb]{0.929,0.969,0.949}}1.77  & {\cellcolor[rgb]{0.929,0.969,0.949}}2.19 & 0.156                                     & {\cellcolor[rgb]{0.988,0.996,0.992}}0.205 & 9.58                                      & 0.682                                      \\
                           & 27B  & Rel.  & {\cellcolor[rgb]{0.733,0.894,0.816}}94.42 & {\cellcolor[rgb]{0.737,0.894,0.82}}1.53  & {\cellcolor[rgb]{0.722,0.886,0.808}}26.03 & {\cellcolor[rgb]{0.749,0.898,0.827}}94.88 & {\cellcolor[rgb]{0.737,0.894,0.816}}4.07 & {\cellcolor[rgb]{0.725,0.886,0.808}}0.293 & {\cellcolor[rgb]{0.729,0.89,0.812}}95.30  & {\cellcolor[rgb]{0.741,0.894,0.82}}1.24   & {\cellcolor[rgb]{0.741,0.894,0.82}}1.50  & {\cellcolor[rgb]{0.757,0.902,0.831}}0.127 & {\cellcolor[rgb]{0.769,0.906,0.839}}0.179 & {\cellcolor[rgb]{0.753,0.902,0.831}}10.65 & {\cellcolor[rgb]{0.761,0.906,0.835}}0.716  \\
                           & 27B  & Abs.  & {\cellcolor[rgb]{0.686,0.875,0.784}}96.46 & {\cellcolor[rgb]{0.698,0.878,0.792}}1.41 & {\cellcolor[rgb]{0.694,0.878,0.792}}26.04 & {\cellcolor[rgb]{0.71,0.882,0.8}}95.15    & {\cellcolor[rgb]{0.675,0.867,0.776}}2.84 & {\cellcolor[rgb]{0.675,0.867,0.773}}0.147 & {\cellcolor[rgb]{0.69,0.875,0.784}}97.03  & {\cellcolor[rgb]{0.698,0.878,0.792}}1.12  & {\cellcolor[rgb]{0.698,0.875,0.792}}1.34 & {\cellcolor[rgb]{0.706,0.882,0.796}}0.121 & {\cellcolor[rgb]{0.71,0.882,0.8}}0.172    & {\cellcolor[rgb]{0.722,0.89,0.808}}10.79  & {\cellcolor[rgb]{0.714,0.886,0.804}}0.722  \\
\midrule
\multirow{2}{*}{E-G (I)}   & 27B  & Rel.  & {\cellcolor[rgb]{0.38,0.749,0.569}}98.90  & {\cellcolor[rgb]{0.361,0.741,0.553}}1.17 & {\cellcolor[rgb]{0.431,0.769,0.604}}26.06 & {\cellcolor[rgb]{0.455,0.78,0.624}}95.69  & {\cellcolor[rgb]{0.506,0.8,0.659}}2.51   & {\cellcolor[rgb]{0.514,0.804,0.663}}0.116 & {\cellcolor[rgb]{0.431,0.773,0.604}}99.26 & {\cellcolor[rgb]{0.341,0.733,0.541}}0.910 & {\cellcolor[rgb]{0.341,0.733,0.541}}1.09 & {\cellcolor[rgb]{0.4,0.757,0.58}}0.109    & {\cellcolor[rgb]{0.435,0.769,0.608}}0.159 & {\cellcolor[rgb]{0.482,0.792,0.639}}11.22 & {\cellcolor[rgb]{0.439,0.773,0.612}}0.736  \\
                           & 27B  & Abs.  & {\cellcolor[rgb]{0.553,0.82,0.69}}97.88   & {\cellcolor[rgb]{0.341,0.733,0.541}}1.16 & {\cellcolor[rgb]{0.498,0.796,0.651}}26.06 & {\cellcolor[rgb]{0.341,0.733,0.541}}95.84 & {\cellcolor[rgb]{0.541,0.812,0.682}}2.55 & {\cellcolor[rgb]{0.388,0.753,0.573}}0.101 & {\cellcolor[rgb]{0.565,0.824,0.698}}98.43 & {\cellcolor[rgb]{0.341,0.733,0.541}}0.910 & {\cellcolor[rgb]{0.341,0.733,0.541}}1.09 & {\cellcolor[rgb]{0.341,0.733,0.541}}0.107 & {\cellcolor[rgb]{0.341,0.733,0.541}}0.156 & {\cellcolor[rgb]{0.353,0.741,0.553}}11.36 & {\cellcolor[rgb]{0.341,0.733,0.541}}0.740  \\
\midrule
\multirow{2}{*}{E-G (T+I)} & 27B  & Rel.  & {\cellcolor[rgb]{0.349,0.737,0.549}}99.08 & {\cellcolor[rgb]{0.451,0.776,0.616}}1.21 & {\cellcolor[rgb]{0.341,0.733,0.541}}26.07 & {\cellcolor[rgb]{0.486,0.792,0.643}}95.65 & {\cellcolor[rgb]{0.482,0.788,0.639}}2.48 & {\cellcolor[rgb]{0.553,0.816,0.69}}0.120  & {\cellcolor[rgb]{0.404,0.761,0.584}}99.44 & {\cellcolor[rgb]{0.443,0.773,0.612}}0.950 & {\cellcolor[rgb]{0.431,0.769,0.604}}1.13 & {\cellcolor[rgb]{0.51,0.8,0.659}}0.112    & {\cellcolor[rgb]{0.529,0.808,0.675}}0.162 & {\cellcolor[rgb]{0.341,0.733,0.541}}11.38 & {\cellcolor[rgb]{0.514,0.804,0.663}}0.734  \\
                           & 27B  & Abs.  & {\cellcolor[rgb]{0.341,0.733,0.541}}99.12 & {\cellcolor[rgb]{0.384,0.749,0.569}}1.18 & {\cellcolor[rgb]{0.404,0.761,0.584}}26.07 & {\cellcolor[rgb]{0.357,0.741,0.553}}95.82 & {\cellcolor[rgb]{0.341,0.733,0.541}}2.32 & {\cellcolor[rgb]{0.341,0.733,0.541}}0.095 & {\cellcolor[rgb]{0.341,0.733,0.541}}99.82 & {\cellcolor[rgb]{0.392,0.753,0.576}}0.930 & {\cellcolor[rgb]{0.361,0.741,0.557}}1.10 & {\cellcolor[rgb]{0.435,0.769,0.604}}0.110 & {\cellcolor[rgb]{0.416,0.761,0.592}}0.158 & {\cellcolor[rgb]{0.365,0.745,0.557}}11.35 & {\cellcolor[rgb]{0.439,0.773,0.612}}0.736  \\
\bottomrule[1pt]
\end{tabular}
}
\end{table*}

\begin{table*}[h]
\centering
\caption{Text-referenced comparisons with general LLMs.}
\resizebox{\linewidth}{!}{%
\begin{tabular}{l|ccccc|cccc|cccc} 
\toprule[1pt]
Model                & {R-ACC}                   & {CD}                     & {CLIP}                  & {DINO}                    & {FID}                   & {FID (C)}               & {R-ACC (U)}                & {CD (T)}                  & {CD (ST)}                & {L2}                    & {LPIPS}                 & {PSNR}                 & {SSIM}                   \\
\midrule
GPT-5 mini           & 4.17                                      & {\cellcolor[rgb]{0.871,0.949,0.91}}10.70 & 24.75                                     & {\cellcolor[rgb]{0.965,0.988,0.976}}79.65 & {\cellcolor[rgb]{0.878,0.949,0.918}}63.86 & 13.78                                     & 4.96                                       & {\cellcolor[rgb]{0.886,0.953,0.922}}10.49 & {\cellcolor[rgb]{0.871,0.949,0.91}}10.74 & 0.382                                     & 0.412                                     & 4.45                                     & 0.432                                      \\
Gemini-2.5 Flash     & {\cellcolor[rgb]{0.996,1,0.996}}4.89      & 13.78                                    & {\cellcolor[rgb]{0.863,0.945,0.906}}25.26 & 78.62                                     & 86.03                                     & {\cellcolor[rgb]{0.965,0.984,0.976}}12.77 & {\cellcolor[rgb]{0.984,0.996,0.992}}7.15   & 13.38                                     & 13.78                                    & {\cellcolor[rgb]{0.945,0.976,0.961}}0.364 & {\cellcolor[rgb]{0.882,0.953,0.918}}0.379 & {\cellcolor[rgb]{0.933,0.973,0.953}}4.85 & {\cellcolor[rgb]{0.882,0.953,0.918}}0.480  \\
Claude Haiku 4.5     & {\cellcolor[rgb]{0.776,0.91,0.847}}32.38  & {\cellcolor[rgb]{0.745,0.894,0.824}}7.60 & {\cellcolor[rgb]{0.769,0.906,0.839}}25.61 & {\cellcolor[rgb]{0.78,0.914,0.847}}84.69  & {\cellcolor[rgb]{0.725,0.886,0.812}}35.07 & {\cellcolor[rgb]{0.749,0.898,0.827}}5.85  & {\cellcolor[rgb]{0.757,0.902,0.831}}38.80  & {\cellcolor[rgb]{0.757,0.902,0.831}}7.17  & {\cellcolor[rgb]{0.741,0.894,0.824}}7.60 & {\cellcolor[rgb]{0.722,0.886,0.808}}0.289 & {\cellcolor[rgb]{0.718,0.882,0.804}}0.332 & {\cellcolor[rgb]{0.733,0.894,0.816}}5.96 & {\cellcolor[rgb]{0.718,0.886,0.804}}0.545  \\
\midrule
GPT-5                & {\cellcolor[rgb]{0.682,0.875,0.78}}43.98  & {\cellcolor[rgb]{0.686,0.871,0.784}}6.12 & {\cellcolor[rgb]{0.678,0.871,0.776}}25.95 & {\cellcolor[rgb]{0.698,0.878,0.792}}86.92 & {\cellcolor[rgb]{0.694,0.875,0.788}}29.00 & {\cellcolor[rgb]{0.682,0.871,0.78}}3.71   & {\cellcolor[rgb]{0.69,0.875,0.788}}48.03   & {\cellcolor[rgb]{0.682,0.871,0.78}}5.26   & {\cellcolor[rgb]{0.686,0.871,0.784}}6.26 & {\cellcolor[rgb]{0.757,0.902,0.831}}0.301 & {\cellcolor[rgb]{0.769,0.906,0.839}}0.346 & {\cellcolor[rgb]{0.784,0.914,0.851}}5.68 & {\cellcolor[rgb]{0.776,0.91,0.847}}0.521   \\
Gemini 2.5 Pro       & {\cellcolor[rgb]{0.843,0.937,0.89}}24.04  & {\cellcolor[rgb]{0.773,0.906,0.843}}8.22 & {\cellcolor[rgb]{0.78,0.914,0.847}}25.57  & {\cellcolor[rgb]{0.812,0.925,0.871}}83.77 & {\cellcolor[rgb]{0.796,0.914,0.859}}47.82 & {\cellcolor[rgb]{0.796,0.918,0.859}}7.39  & {\cellcolor[rgb]{0.843,0.937,0.89}}27.06   & {\cellcolor[rgb]{0.776,0.91,0.847}}7.72   & {\cellcolor[rgb]{0.769,0.906,0.839}}8.22 & {\cellcolor[rgb]{0.812,0.922,0.871}}0.319 & {\cellcolor[rgb]{0.769,0.906,0.839}}0.347 & {\cellcolor[rgb]{0.816,0.925,0.875}}5.48 & {\cellcolor[rgb]{0.784,0.914,0.851}}0.518  \\
Claude Sonnet 4.5    & {\cellcolor[rgb]{0.663,0.867,0.769}}46.65 & {\cellcolor[rgb]{0.635,0.851,0.745}}5.28 & {\cellcolor[rgb]{0.663,0.863,0.769}}25.99 & {\cellcolor[rgb]{0.639,0.855,0.749}}88.31 & {\cellcolor[rgb]{0.596,0.835,0.718}}19.59 & {\cellcolor[rgb]{0.62,0.847,0.737}}2.81   & {\cellcolor[rgb]{0.655,0.863,0.765}}52.99  & {\cellcolor[rgb]{0.635,0.851,0.749}}4.58  & {\cellcolor[rgb]{0.631,0.851,0.745}}5.34 & {\cellcolor[rgb]{0.624,0.847,0.737}}0.253 & {\cellcolor[rgb]{0.631,0.851,0.745}}0.305 & {\cellcolor[rgb]{0.643,0.855,0.753}}6.62 & {\cellcolor[rgb]{0.631,0.851,0.745}}0.580  \\
\midrule
VecGlypher 27B T,I,R & {\cellcolor[rgb]{0.349,0.737,0.545}}99.62 & {\cellcolor[rgb]{0.345,0.733,0.545}}1.77 & {\cellcolor[rgb]{0.349,0.737,0.549}}26.53 & {\cellcolor[rgb]{0.357,0.741,0.553}}93.97 & {\cellcolor[rgb]{0.341,0.733,0.541}}3.50  & {\cellcolor[rgb]{0.341,0.733,0.541}}0.139 & {\cellcolor[rgb]{0.349,0.737,0.549}}99.58  & {\cellcolor[rgb]{0.349,0.737,0.545}}1.51  & {\cellcolor[rgb]{0.349,0.733,0.545}}1.76 & {\cellcolor[rgb]{0.349,0.737,0.545}}0.143 & {\cellcolor[rgb]{0.353,0.737,0.549}}0.197 & {\cellcolor[rgb]{0.361,0.741,0.557}}9.64 & {\cellcolor[rgb]{0.353,0.737,0.549}}0.695  \\
VecGlypher 27B T,I,A & {\cellcolor[rgb]{0.341,0.733,0.541}}100.5 & {\cellcolor[rgb]{0.341,0.733,0.541}}1.72 & {\cellcolor[rgb]{0.349,0.737,0.549}}26.53 & {\cellcolor[rgb]{0.345,0.737,0.545}}94.22 & {\cellcolor[rgb]{0.341,0.733,0.541}}3.46  & {\cellcolor[rgb]{0.341,0.733,0.541}}0.134 & {\cellcolor[rgb]{0.345,0.737,0.545}}100.34 & {\cellcolor[rgb]{0.341,0.733,0.541}}1.44  & {\cellcolor[rgb]{0.341,0.733,0.541}}1.68 & {\cellcolor[rgb]{0.341,0.733,0.541}}0.140 & {\cellcolor[rgb]{0.341,0.733,0.541}}0.193 & {\cellcolor[rgb]{0.345,0.737,0.545}}9.83 & {\cellcolor[rgb]{0.345,0.737,0.545}}0.698  \\
\midrule
VecGlypher 70B T,R   & {\cellcolor[rgb]{0.345,0.737,0.545}}100.1 & {\cellcolor[rgb]{0.341,0.733,0.541}}1.70 & {\cellcolor[rgb]{0.349,0.737,0.545}}26.53 & {\cellcolor[rgb]{0.353,0.737,0.549}}94.10 & {\cellcolor[rgb]{0.341,0.733,0.541}}3.45  & {\cellcolor[rgb]{0.341,0.733,0.541}}0.140 & {\cellcolor[rgb]{0.345,0.737,0.545}}100.57 & {\cellcolor[rgb]{0.341,0.733,0.541}}1.43  & {\cellcolor[rgb]{0.341,0.733,0.541}}1.67 & {\cellcolor[rgb]{0.341,0.733,0.541}}0.140 & {\cellcolor[rgb]{0.341,0.733,0.541}}0.193 & {\cellcolor[rgb]{0.349,0.737,0.545}}9.80 & {\cellcolor[rgb]{0.341,0.733,0.541}}0.699  \\
VecGlypher 70B T,A   & {\cellcolor[rgb]{0.345,0.737,0.545}}100.4 & {\cellcolor[rgb]{0.341,0.733,0.541}}1.68 & {\cellcolor[rgb]{0.341,0.733,0.541}}26.54 & {\cellcolor[rgb]{0.341,0.733,0.541}}94.28 & {\cellcolor[rgb]{0.341,0.733,0.541}}3.34  & {\cellcolor[rgb]{0.341,0.733,0.541}}0.136 & {\cellcolor[rgb]{0.341,0.733,0.541}}100.71 & {\cellcolor[rgb]{0.341,0.733,0.541}}1.40  & {\cellcolor[rgb]{0.341,0.733,0.541}}1.66 & {\cellcolor[rgb]{0.341,0.733,0.541}}0.139 & {\cellcolor[rgb]{0.341,0.733,0.541}}0.192 & {\cellcolor[rgb]{0.341,0.733,0.541}}9.85 & {\cellcolor[rgb]{0.341,0.733,0.541}}0.699  \\
\bottomrule[1pt]
\end{tabular}
}
\end{table*}

\begin{table*}[h]
\centering
\caption{Image-referenced comparisons with vector-font baselines.}
\resizebox{\linewidth}{!}{%
\begin{tabular}{l|ccccc|cccc|cccc} 
\toprule[1pt]
Model                & {R-ACC}                   & {CD}                      & {CLIP}                  & {DINO}                    & {FID}                   & {FID (C)}               & {R-ACC (U)}               & {CD (T)}                  & {CD (ST)}                 & {L2}                    & {LPIPS}                 & {PSNR}                  & {SSIM}                   \\
\midrule
DeepVecFont-v2       & 37.86                                     & {\cellcolor[rgb]{0.925,0.969,0.949}}14.58 & 24.81                                     & 79.41                                     & 115.5                                     & 16.45                                     & 49.29                                     & {\cellcolor[rgb]{0.929,0.969,0.949}}14.58 & {\cellcolor[rgb]{0.925,0.969,0.949}}14.58 & 0.243                                     & 0.320                                     & 6.70                                      & 0.599                                      \\
DualVector           & {\cellcolor[rgb]{0.898,0.961,0.929}}49.20 & 16.45                                     & {\cellcolor[rgb]{0.89,0.957,0.925}}25.07  & {\cellcolor[rgb]{0.996,1,0.996}}79.57     & {\cellcolor[rgb]{0.945,0.976,0.961}}105.5 & {\cellcolor[rgb]{0.933,0.973,0.953}}14.73 & {\cellcolor[rgb]{0.839,0.937,0.89}}65.59  & 16.44                                     & 16.44                                     & {\cellcolor[rgb]{0.808,0.922,0.867}}0.190 & {\cellcolor[rgb]{0.902,0.961,0.933}}0.293 & {\cellcolor[rgb]{0.859,0.945,0.902}}7.99  & {\cellcolor[rgb]{0.812,0.925,0.871}}0.653  \\
\midrule
VecGlypher 27B T,I,R & {\cellcolor[rgb]{0.345,0.737,0.545}}99.08 & {\cellcolor[rgb]{0.341,0.733,0.541}}1.21  & {\cellcolor[rgb]{0.341,0.733,0.541}}26.07 & {\cellcolor[rgb]{0.349,0.737,0.549}}95.65 & {\cellcolor[rgb]{0.341,0.733,0.541}}2.48  & {\cellcolor[rgb]{0.341,0.733,0.541}}0.120 & {\cellcolor[rgb]{0.349,0.737,0.549}}99.44 & {\cellcolor[rgb]{0.341,0.733,0.541}}0.950 & {\cellcolor[rgb]{0.341,0.733,0.541}}1.13  & {\cellcolor[rgb]{0.357,0.737,0.553}}0.112 & {\cellcolor[rgb]{0.357,0.737,0.553}}0.162 & {\cellcolor[rgb]{0.341,0.733,0.541}}11.38 & {\cellcolor[rgb]{0.365,0.745,0.557}}0.734  \\
VecGlypher 27B T,I,R & {\cellcolor[rgb]{0.341,0.733,0.541}}99.12 & {\cellcolor[rgb]{0.341,0.733,0.541}}1.18  & {\cellcolor[rgb]{0.345,0.737,0.545}}26.07 & {\cellcolor[rgb]{0.341,0.733,0.541}}95.82 & {\cellcolor[rgb]{0.341,0.733,0.541}}2.32  & {\cellcolor[rgb]{0.341,0.733,0.541}}0.095 & {\cellcolor[rgb]{0.341,0.733,0.541}}99.82 & {\cellcolor[rgb]{0.341,0.733,0.541}}0.930 & {\cellcolor[rgb]{0.341,0.733,0.541}}1.10  & {\cellcolor[rgb]{0.341,0.733,0.541}}0.110 & {\cellcolor[rgb]{0.341,0.733,0.541}}0.158 & {\cellcolor[rgb]{0.349,0.737,0.545}}11.35 & {\cellcolor[rgb]{0.341,0.733,0.541}}0.736  \\
\bottomrule[1pt]
\end{tabular}
}
\end{table*}

\begin{table*}[h]
\centering
\caption{Performance of the open weight LLMs and the vector graphic LLM.}
\resizebox{\linewidth}{!}{%
\begin{tabular}{l|ccccc|cccc|cccc} 
\toprule[1pt]
Model                       & {R-ACC}                  & {CD}                      & {CLIP}                    & {DINO}                    & {FID}                      & {FID (C)}                & {R-ACC (U)}              & {CD (T)}                  & {CD (ST)}                 & {L2}                    & {LPIPS}                 & {PSNR}                  & {SSIM}                   \\
\midrule
Llama3.3 70B Instruct       & {\cellcolor[rgb]{0.808,0.925,0.871}}0.08 & {\cellcolor[rgb]{0.776,0.906,0.843}}38.28 & {\cellcolor[rgb]{0.663,0.863,0.765}}24.84  & {\cellcolor[rgb]{0.655,0.863,0.761}}75.85 & {\cellcolor[rgb]{0.667,0.863,0.769}}143.73 & {\cellcolor[rgb]{0.671,0.867,0.773}}21.49 & {\cellcolor[rgb]{0.875,0.949,0.914}}0.08 & {\cellcolor[rgb]{0.776,0.91,0.843}}38.26  & {\cellcolor[rgb]{0.776,0.91,0.843}}38.30  & {\cellcolor[rgb]{0.643,0.855,0.753}}0.374 & {\cellcolor[rgb]{0.655,0.859,0.761}}0.388 & {\cellcolor[rgb]{0.659,0.863,0.765}}4.64  & {\cellcolor[rgb]{0.639,0.855,0.753}}0.480  \\
Gemma3 27B IT               & {\cellcolor[rgb]{0.698,0.878,0.792}}0.12 & {\cellcolor[rgb]{0.557,0.82,0.694}}17.90  & {\cellcolor[rgb]{0.671,0.867,0.773}}24.78 & {\cellcolor[rgb]{0.702,0.882,0.796}}74.26 & {\cellcolor[rgb]{0.737,0.894,0.82}}162.89  & {\cellcolor[rgb]{0.71,0.882,0.8}}23.24    & {\cellcolor[rgb]{0.82,0.929,0.875}}0.11  & {\cellcolor[rgb]{0.553,0.82,0.69}}17.44   & {\cellcolor[rgb]{0.557,0.82,0.694}}17.90  & {\cellcolor[rgb]{0.71,0.882,0.796}}0.407  & {\cellcolor[rgb]{0.675,0.867,0.776}}0.397 & {\cellcolor[rgb]{0.722,0.89,0.808}}4.28 & {\cellcolor[rgb]{0.69,0.878,0.788}}0.450   \\
Qwen3 30B A3B Instruct 2507 & {\cellcolor[rgb]{0.671,0.867,0.773}}0.39 & {\cellcolor[rgb]{0.671,0.867,0.773}}26.15 & {\cellcolor[rgb]{0.698,0.878,0.792}}24.66 & {\cellcolor[rgb]{0.718,0.886,0.804}}74.03 & {\cellcolor[rgb]{0.682,0.871,0.78}}148.78  & {\cellcolor[rgb]{0.69,0.875,0.788}}22.47  & {\cellcolor[rgb]{0.671,0.867,0.773}}0.55 & {\cellcolor[rgb]{0.671,0.867,0.773}}26.07 & {\cellcolor[rgb]{0.671,0.867,0.773}}26.15 & {\cellcolor[rgb]{0.741,0.894,0.82}}0.418  & {\cellcolor[rgb]{0.82,0.925,0.875}}0.422  & {\cellcolor[rgb]{0.741,0.898,0.824}}4.219 & {\cellcolor[rgb]{0.769,0.906,0.839}}0.427  \\
\midrule
OmniSVG                     & 0.01                                     & 63.70                                     & 23.26                                     & 69.35                                     & 229.38                                     & 35.75                                     & 0.01                                     & 63.69                                     & 63.69                                     & 0.496                                     & 0.452                                     & 3.50                                    & 0.356                                      \\
\bottomrule[1pt]
\end{tabular}
}
\end{table*}

\end{document}